\documentclass[11pt]{article}

\usepackage[]{acl}

\usepackage{times}
\usepackage{latexsym}

\usepackage[T1]{fontenc}
\usepackage[utf8]{inputenc}

\usepackage{microtype}

\usepackage{inconsolata}

\usepackage{graphicx}
\usepackage{multirow}
\usepackage{CJKutf8}
\usepackage[utf8]{inputenc}
\usepackage{booktabs}
\usepackage{tabularx}
\usepackage{array}
\usepackage{makecell}
\usepackage{amssymb}

\usepackage{tcolorbox}
\tcbuselibrary{breakable, skins}

\newtcolorbox{promptbox}[1]{
  breakable,                    % allows the box to split across pages
  colback=gray!4,               % very light gray background
  colframe=gray!55,             % medium gray border
  coltitle=black,               % title text color
  fonttitle=\bfseries\small,    % title font
  fontupper=\small,             % body font size
  title=#1,                     % the title is passed as an argument
  boxrule=0.5pt,                % border thickness
  left=5pt, right=5pt,          % inner horizontal padding
  top=4pt, bottom=4pt,          % inner vertical padding
  enhanced,                     % needed for some skin features
  attach boxed title to top left={xshift=6pt, yshift=-2pt},
  boxed title style={
    colback=gray!20,
    colframe=gray!55,
    boxrule=0.5pt,
  },
}
\title{\textsc{CNeo-Bench}: Diagnosing Large Language Models on \\ 
Chinese Neologisms}

\author{
 \textbf{Kaiyan Zhao\textsuperscript{1}},
 \textbf{Zhongtao Miao\textsuperscript{1}},
 \textbf{Zheyong Xie\textsuperscript{2}}, \\
 \textbf{Shaosheng Cao\textsuperscript{2,3},
 \textbf{Yoshimasa Tsuruoka\textsuperscript{1}}}
\\
 \textsuperscript{1}The University of Tokyo,
 \textsuperscript{2}Xiaohongshu Inc.,
 \textsuperscript{3}Tsinghua University
\\
\normalfont{\fontsize{11pt}{12pt}\selectfont {\fontfamily{qcr}\selectfont kaiyan1006@logos.t.u-tokyo.ac.jp}} \\
}

\begin{document}
\begin{CJK*}{UTF8}{gbsn}
\maketitle
\begin{abstract}
Chinese neologisms exploit diverse and unique linguistic mechanisms, such as phonetic substitution (e.g., \textit{886} for ``bye-bye'') and visual character decomposition (e.g., \textit{彳亍}~for~\textit{行}, ``okay'') that are rare in other languages. We introduce \textsc{CNeo-Bench}, a benchmark of 4{,}759 such neologisms with reference definitions, organized into five top-level categories and nine subcategories by the linguistic mechanism behind each expression. 
\textsc{CNeo-Bench} is paired with a two-tier evaluation framework 
that separates whether a model can \textit{describe} a neologism 
from whether it can \textit{operate on} its underlying mechanism. 
Evaluating 18 LLMs, we find that Chinese neologisms remain an 
open challenge; most models fall below 40\% on definition generation, and on several subcategories a systematic 
\textit{recognition-manipulation gap} emerges: models describe 
neologisms correctly but, in source-form restoration tasks, substitute a 
semantic equivalent (paraphrase) for the source form rather 
than producing the source form itself. A few-shot analysis on 1{,}058 hard items shows that in-context examples can solve many difficult cases, but leave a noticeable portion of errors remaining, indicating 
challenges beyond prompting alone can address\footnote{Data and code will be released upon acceptance.}.
\end{abstract}

\section{Introduction}
Neologisms\footnote{\url{https://en.wikipedia.org/wiki/Neologism}}, newly coined or re-combined expressions that emerge from everyday human communication, have been an arising research target for Natural Language Processing~(NLP)~\citep{mccrae-2019-identification-neologisms, zalmout-etal-2019-unsupervised-neologism-norm, zheng-etal-2024-neobench}, accelerated by the rapid creation and diffusion of new expressions on social media~\citep{zhao2026benchmarkingmachinetranslationchinese}.
How a Large Language Model~(LLM) handles neologisms is a useful probe of its linguistic capability, since these expressions lie at or beyond the edge of any model's training distribution.

While neologisms exist in every living language, Chinese neologisms pose a particularly distinctive challenge~\citep{CanLargeLanguageModelsUnderstandChineseNeologisms, miao2026neoamt}. Beyond the lexical neologisms~(new words) and the semantic neologisms~(existing words convey new meanings) that can be observed in most languages, Chinese neologisms systematically exploit more diverse linguistic mechanisms. For example, 
phonetic substitution between digits and Mandarin phrases (``886'' pronounced like ``拜拜咯'', ``bye-bye''), visual decomposition of characters (``彳亍'' decomposed from ``行'', okay''), and cross-script abbreviation from \textit{Pinyin}\footnote{The romanization system for Mandarin Chinese. See \url{https://en.wikipedia.org/wiki/Pinyin}.} (``yyds'' from \textbf{y}ong \textbf{y}uan \textbf{d}e \textbf{s}hen'', ``forever legendary''). The evaluation of these mechanisms requires tasks designed around the specific linguistic operations that produce them~\citep{CanLargeLanguageModelsUnderstandChineseNeologisms}.

\begin{figure}[t]
  \centering
  \includegraphics[width=0.49\textwidth]{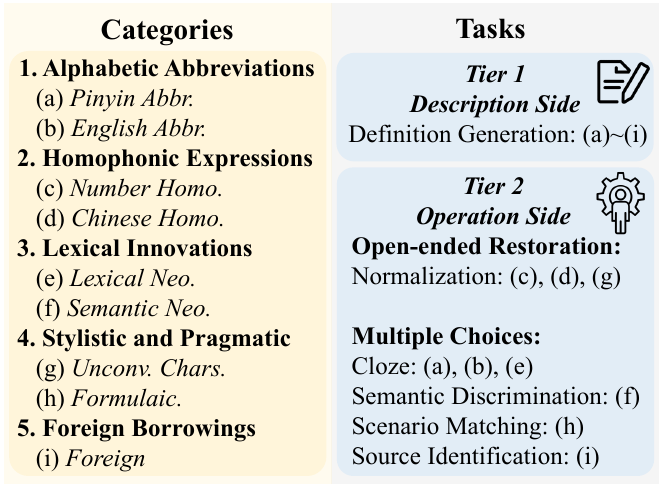}
  \caption{Categories and tasks in \textsc{CNeo-Bench}.}
  \label{category_figure}

\end{figure}
To this end, we introduce \textsc{CNeo-Bench}, a benchmark of 4{,}759 Chinese neologisms, each paired with a reference definition, organized by the linguistic mechanism behind each expression, spanning five top-level categories and nine subcategories as shown in Figure~\ref{category_figure}. The benchmark is paired with a \textbf{two-tier evaluation framework} that distinguishes two capabilities models may have to different degrees: a \textit{definition generation task} shared across all subcategories, which measures whether a model can \textit{describe} a neologism; and \textit{category-specific diagnostic tasks} targeting each mechanism, which measure whether a model can \textit{operate on} the form behind it. This distinction matters because the two do not coincide: a model may correctly describe ``彳亍口巴'' in Tier 1 as the visually decomposed form of ``行吧'' (okay), yet in the Tier 2 task, the same model produces paraphrase rather than the source form\footnote{A failure case from DeepSeek-V3.2, more cases are provided in Table~\ref{tab:case-study}, Appendix.}.

We evaluate 18 LLMs spanning multiple families and parameter scales on \textsc{CNeo-Bench}. We find that Chinese neologisms remain an open challenge even for frontier models, with the strongest, Kimi-K2.5~\citep{kimiteam2026kimik25visualagentic}, reaches only 67.74\% on definition generation. More importantly, the two-tier framework surfaces patterns that aggregate scoring would hide. On subcategories requiring open-ended restoration of a source form, we observe a systematic \textbf{recognition-manipulation gap}: models describe these neologisms correctly but fail to produce the source form itself, instead copying the neologism verbatim or producing a semantic equivalent. We additionally introduce an analysis on 1{,}058 hard samples (joint failure cases for three frontier models), and show that providing a few in-context examples recovers a substantial portion. A persistent residual resists in-context recovery even at three shots, pointing to challenges that contextual grounding alone cannot address.

Our contributions are as follows:
\begin{itemize}
    \item We introduce \textsc{CNeo-Bench}, a benchmark of 4{,}759 Chinese neologisms paired with reference definition, organized into five top-level categories and nine subcategories by linguistic formation mechanism, and a \textbf{two-tier evaluation framework} that separates whether a model can describe a neologism from whether it can operate on its source form.
    
    \item Evaluating 18 LLMs, we identify a systematic \textbf{recognition-manipulation gap} on the specific subcategories: models describe Chinese neologisms correctly but fail to produce their source forms, instead substituting a semantic equivalent for the source form, a pattern that persists across models.
    
    \item A few-shot analysis on hard items shows that in-context examples can solve difficult cases, but a noticeable portion of errors remains.
    
\end{itemize}

\section{\textsc{Cneo-Bench}}
\subsection{Taxonomy}
\label{sec:taxonomy}
\textsc{CNeo-Bench} organizes neologisms by the linguistic mechanism through which they are formed.
%, rather than by topical domain or surface frequency. 
% This mechanism-driven grouping lets us diagnose where model failures actually occur, linking each observed failure to a specific formation process.
Our taxonomy extends the folk classification of Chinese internet language on Wikiversity~\citep{wikiversity-neologisms}, refined through reference with prior work on Chinese neologisms~\citep{CanLargeLanguageModelsUnderstandChineseNeologisms}. We introduce a two-level structure with five top-level categories and nine subcategories, with an overview in Figure~\ref{category_figure}, detailed taxonomy in Appendix~\ref{detailed_tax}, and examples in Table~\ref{category_examples}.

\paragraph{(1) Alphabetic Abbreviations.} Initialism-style forms 
abbreviating a phrase into Latin letters: \textit{Pinyin Abbr.} 
takes initials from the Pinyin of a Chinese phrase (\textit{yyds} 
← 永远的神, ``forever legendary''), while \textit{English Abbr.} 
takes initials from an English phrase used in Chinese contexts 
(\textit{ACG} ← \textit{Anime, Comics, Games}).

\paragraph{(2) Homophonic Expressions.} Forms encoding a target 
phrase through phonetic substitution. \textit{Number Homo.} uses 
numbers read aloud as the target phrase (\textit{886} ← 
拜拜咯, ``bye-bye''); \textit{Chinese Homo.} substitutes characters 
of the target with phonetically similar but semantically 
unrelated ones (雨女无瓜 ← 与你无关, ``none of your business'').

\paragraph{(3) Lexical Innovations.} Forms whose novelty is at the 
lexical-semantic level. \textit{Lexical Neo.} introduces new items (键盘侠, ``keyboard warrior''); \textit{Semantic Neo.} 
shifts an existing word to a new meaning (养鱼, literally ``fish 
farming'', now used for maintaining multiple romantic prospects).

\paragraph{(4) Stylistic and Pragmatic Expressions.} Forms fixed 
by convention rather than productive rules. \textit{Unconv. Chars.} 
decomposes a standard character into its visual components written 
separately (彳亍 ← 行, ``okay''); \textit{Formulaic} expressions 
are full phrases reused as units, typically originating from viral 
media (没活了可以咬打火机, mocking someone out of fresh content).

\paragraph{(5) Foreign Borrowings.} Forms derived from foreign 
sources through multiple paths: phonetic reinterpretation 
(逮虾户 ← \textit{Déjà Vu}) and direct adoption from 
other languages (文化祭, ``cultural festival'', from Japanese). These paths are treated 
as a single subcategory because they share the defining feature 
of dependence on a non-Chinese source.

\subsection{Data Collection}
\textsc{CNeo-Bench} contains 4,759 Chinese neologisms, each paired with a reference definition. All data were collected through January 2026. We describe the collection pipeline, annotation protocol, and evaluation-time quality controls below.

\paragraph{Source discovery.} We begin with exploratory Google Search API queries using manually designed templates targeting Chinese neologisms (e.g., ``\rule{0.5cm}{0.5pt}， 这是什么梗'', what does this 梗~mean; ``\rule{0.5cm}{0.5pt}，这个梗是什么意思'', what is the meaning of this 梗\footnote{In Chinese, the term ``梗'' is the more commonly used 
term for what the NLP literature calls ``neologism''.}). This initial pool of approximately 50,000 snippets is noisy, with a substantial fraction consisting of unrelated advertisements or content not tied to any specific neologism. However, inspecting the pool reveals that high-quality, well-defined neologism entries cluster heavily on a small number of community-maintained sources, e.g., meme wikis and dictionaries. We use this observation to guide targeted crawling in the next stage.

\paragraph{Targeted crawling.} For each identified source, we check the licensing terms and proceed only with sources whose licenses permit research use, mostly CC BY-NC-SA 3.0. Moegirlpedia\footnote{\url{https://zh.moegirl.org.cn}} serves as the largest contributor, while Wikiversity's list of Chinese mainland internet slang~\citep{wikiversity-neologisms} is the second major source, from which we retrieve 1,964 and 1,588 entries individually. The remaining entries are drawn from additional websites or dictionaries acquired through the initial Google Search API pool. For all entries, we extract only the neologism, its reference definition, and example usage sentences if exist. 

\paragraph{Consolidation and filtering.} Entries from all sources are deduplicated and then merged. We then perform a machine-verification pass to confirm that each entry (i) is paired with a coherent definition, and (ii) does not contain sensitive or harmful content. This pass yields 4,917 entries. To further exclude rare items, we apply a popularity filter using RedNote, a major Chinese social platform. Specifically, we use the neologism as a keyword and require at least 5 returned posts for the entry to be retained. This removed 158 entries with negligible real-world usage, leaving the final dataset of 4,759 entries. The source distribution of the final dataset is: Moegirlpedia (1,393 entries, 29.3\%), Wikiversity (1,024 entries, 21.5\%), and other web sources via Google Search (2,342 entries, 49.2\%).

\paragraph{Annotation.} Each of the 4,759 entries is independently labeled with one of the nine subcategories in our taxonomy (Section~\ref{sec:taxonomy}) by three part-time annotators. 
Raw three-way agreement on the 9-category scheme is 78\% (Fleiss' $\kappa$=0.69). Details are provided in Appendix~\ref{annotation}. The statistics for \textsc{CNeo-Bench} is provided in Table~\ref{statistics}.
\begin{table}[t]
\centering
\small
\setlength{\tabcolsep}{6pt}
\renewcommand{\arraystretch}{1.1}
\begin{tabular}{l c c c}
\toprule
\textbf{Subcategory} & \textbf{Full} & \textbf{Hard} & \textbf{Hard Rate} \\
\midrule
Pinyin Abbr.   &  192 &   35 & 18.23\% \\
English Abbr.  &   58 &    7 & 12.07\% \\
Chinese Homo.   &  377 &   65 & 17.24\% \\
Number Homo.    &   29 &    7 & 24.14\% \\
Lexical Neo.         & 1161 &  298 & 25.67\% \\
Semantic Neo.        &  850 &  185 & 21.76\% \\
Unconv. Char.   &   23 &    3 & 13.04\% \\
Formulaic.       & 1667 &  347 & 20.82\% \\
Foreign         &  402 &  111 & 27.61\% \\
\midrule
\textbf{Total}  & \textbf{4759} & \textbf{1058} & \textbf{22.23\%} \\
\bottomrule
\end{tabular}
\caption{Statistics for \textsc{CNeo-Bench}. 
``Full'' is the total number of items in each subcategory; 
``Hard'' is the number of items on which GPT-5.1, DeepSeek-V3.2, and Kimi-K2.5 jointly fail under 
definition generation.}
\label{statistics}
\end{table}

\subsection{Task Design}
\textsc{CNeo-Bench} evaluates models through a two-tier architecture designed to distinguish two distinct capabilities: \textit{(1) Description Side}, to test whether a model knows what a specific neologism means, through a definition generation task shared across all subcategories, and \textit{(2) Operation Side}, to measure whether a model can operate on the neologism correctly beyond merely generating a definition, through subcategory-specific diagnostic tasks.

\paragraph{Tier 1: Definition Generation.} Given a neologism, the model generates a 0-shot definition describing its meaning and usage. This task is applied uniformly to all entries. Model outputs are scored against reference definitions using an LLM-based judge~\citep{gu2024surveyllmasjudge}.

\paragraph{Tier 2: Category-Specific Diagnostics.} Tier 2 tasks take one of two forms, depending on whether the subcategory admits a unique decodable source form. For three subcategories, we use \textbf{open-ended restoration}, where the model produces the source form directly and outputs are evaluated by exact string match. For \textit{Unconv. Chars.}, the neologism alone is given as input, since decomposition is deterministic (彳亍 → 行). For \textit{Chinese Homo.} and \textit{Number Homo.}, a given neologism may plausibly map to multiple standard phrases, so the model is given an example context along with the neologism to produce the source form (雨女无瓜 → 与你无关; 886 → 拜拜咯). For the remaining six subcategories, we use \textbf{multiple-choice (MC)} with four options, with the specific formulations below:

\begin{itemize}
\item \textit{Pinyin Abbr.}, \textit{English Abbr.} and \textit{Lexical Neo.}: cloze. A sentence containing the neologism is presented with the target word masked, and the model selects the correct neologism from four candidates.

\item \textit{Semantic Neo.}: semantic discrimination. The model is presented with four sentences, each containing the same target word. Three of them use the standard meaning and one uses the new meaning. The model is prompted to select the sentence using the new meaning.

\item \textit{Formulaic.}: scenario matching. Given a descriptive scenario, the model selects which of four candidate formulaic expressions best fits the pragmatic context. Distractors are drawn from the gold answers of other items in the same subcategory, randomized for each item, such that only 
a model recognizing the conventionalized meaning of the target 
expression can reliably select the correct answer over other 
plausible expressions of the same type.

\item \textit{Foreign}: source identification. The model selects which of the four options correctly identifies the foreign source and the borrowing mechanism of a given neologism. Distractors are constructed to be structurally parallel as the answer, differing only at the specific details such as source language or borrowing path that a model must know to answer.

\end{itemize}

Table~\ref{tab:task-examples}, Appendix, provides examples for Tier 2 tasks.
Several Tier 2 tasks require example sentences or scenarios 
containing the target neologism, and the multiple-choice tasks 
require distractors. We construct both using an LLM and verify 
them manually, using original example sentences from source 
materials where available (492 entries). Prompts we use are provided in 
Appendix~\ref{app:prompts}.
% Several Tier 2 tasks require example sentences or scenarios containing the target neologism. Where original example sentences are available (492 entries), we use them directly. For the remaining entries, we generated candidate sentences using an LLM prompted with the neologism and its verified definition, with the prompt explicitly instructing the model not to paraphrase the definition within the sentence. Generated sentences were manually reviewed to ensure natural and correct usage of the target neologism. Similarly, for MC tasks requiring distractors, distractors were initially produced by an LLM and subsequently refined by the authors to improve distractor quality and diversity. Full details of prompt templates, the distractor construction procedure for each task, and example task instances are provided in Appendix~\ref{app:prompts}.

\begin{table*}[ht]
\centering
\small
\setlength{\tabcolsep}{5pt}
\renewcommand{\arraystretch}{1.15}
\resizebox{0.95\textwidth}{!}{
\begin{tabular}{l cc cc cc cc c c}
\toprule
 \multirow{2}{*}{\textbf{Model}} & \multicolumn{2}{c}{\textbf{Alphabetic}} & \multicolumn{2}{c}{\textbf{Homophonic}} & \multicolumn{2}{c}{\textbf{Lexical}} & \multicolumn{2}{c}{\textbf{Stylistics}} & \multirow{2}{*}{\textbf{Foreign}} & \multirow{2}{*}{\textbf{Avg.$^\ast$}} \\
\cmidrule(lr){2-3} \cmidrule(lr){4-5} \cmidrule(lr){6-7} \cmidrule(lr){8-9}
 & Pinyin & English & Chinese & Number & Lexical & Semantic & Unconv. Char. & Formulaic & \\
\midrule
GPT-5.1                 & 47.40 & \underline{74.14} & 48.81 & \underline{44.83} & 49.10 & 54.94 & 21.74 & 52.67 & 42.54 & 50.89 \\
\addlinespace
DeepSeek-V3             & 64.06 & \underline{74.14} & 55.44 & \textbf{72.41} & 51.51 & 54.59 & 65.22 & 54.35 & 45.27 & 53.81 \\
DeepSeek-V3.2           & 58.85 & \textbf{79.31} & 60.21 & \textbf{72.41} & 54.61 & 55.65 & \underline{69.57} & 54.17 & 49.00 & 55.26 \\
\addlinespace
Kimi-K2-Instruct        & \underline{66.15} & 70.69 & \underline{61.27} & \textbf{72.41} & \underline{58.14} & \underline{64.47} & 65.22 & \underline{62.57} & \underline{53.48} & \underline{61.27} \\
Kimi-K2.5               & \textbf{76.56} & \textbf{79.31} & \textbf{70.82} & \textbf{72.41} & \textbf{66.06} & \textbf{67.06} & \textbf{82.61} & \textbf{67.55} & \textbf{61.69} & \textbf{67.74} \\
\midrule
Qwen3-4B                & 9.38 & 41.38 & 15.53 & 17.24 & 24.89 & 28.12 & 8.70 & 26.03 & 13.93 & 23.49\\
Qwen3-4B-Instruct-2507  & 7.81 & 53.45 & 16.71 & 20.69 & 27.82 & 32.94 & 13.04 & 28.91 & 16.67 & 26.69\\
Qwen3-8B                & 17.71 & 53.45 & 18.04 & 27.59 & 31.35 & 33.65 & 4.55 & 32.03  & 18.16 & 29.40\\
Qwen3-32B               & \underline{21.88} & \textbf{63.79} & \underline{25.99} & \underline{34.48} & \underline{36.69} & \underline{39.18} & \underline{17.39} & \underline{37.79} & \underline{25.37} & \underline{35.34} \\
\addlinespace
GLM-4-9B-0414           & 18.75 & 48.28 & 20.16 & 20.69 & 29.46 & 34.00 & 8.70 & 29.03 & 21.39 & 28.35\\
GLM-4-32B-0414          & \textbf{30.21} & \textbf{63.79} & \textbf{35.28} & \textbf{65.52} & \textbf{41.26} & \textbf{45.88} & \textbf{34.78} & \textbf{41.0}9 & \textbf{30.60} & \textbf{40.60}\\
\addlinespace
InternLM3-8B-Instruct   & 17.19 & \underline{58.62} & 18.30 & \underline{34.48} & 25.84 & 31.06 & 4.55 & 20.94 & 15.17 & 23.56\\
\addlinespace
% \midrule
Gemma3-4B-it            & 5.21 & 37.93 & 7.69 & 3.45 & 17.23 & 19.29 & 0.00 & 16.68 & 10.45 & 15.68\\
Gemma3-12B-it           & 19.27 & 48.28 & 14.32 & 17.24 & 25.58 & 27.06 & 4.55 & 23.34 & 18.16 & 23.41\\
Gemma3-27B-it           & 26.04 & \textbf{63.79} & 21.22 & 13.79 & 31.52 & 30.00 & 4.55 & 29.57 & 19.40 & 28.66 \\
\addlinespace
Llama3.2-1B-Instruct    & 1.56 & 3.45 & 3.18 & 10.34 & 5.77 & 4.47 & 0.00 & 6.24  & 2.24 & 5.00\\
Llama3.2-3B-Instruct    & 5.21 & 24.14 & 3.71 & 3.45 & 5.77 & 8.71 & 0.00 & 8.64  & 6.22 & 7.33\\
Llama3.2-11B-Instruct   & 4.69 & 34.48 & 7.16 & 3.45 & 14.13 & 17.29 & 0.00 & 13.44  & 10.70 & 13.34\\
\bottomrule
\end{tabular}}
\caption{Full definition generation task results. 
Scores are reported as accuracy (\%). 
\textbf{Bold} indicates the best score in each column; \underline{underline} indicates the second best. Avg.$^\ast$ is computed as micro-average across all 4,759 instances.}
\label{definition_full}
\end{table*}

\section{Experiments}

\subsection{Evaluated Models}
We evaluate 18 LLMs on \textsc{Cneo-Bench}, including GPT-5.1~\citep{singh2025openaigpt5card}; DeepSeek-V3~\citep{deepseekai2025deepseekv3technicalreport} and V3.2~\citep{deepseekai2025deepseekv32pushingfrontieropen}; Kimi-K2-Instruct~\citep{kimiteam2026kimik2openagentic} and Kimi-K2.5~\citep{kimiteam2026kimik25visualagentic}; the Qwen3 series (4B, 4B-Instruct-2507, 8B and 32B)~\citep{yang2025qwen3technicalreport}; the GLM-4 series (9B, 32B)~\citep{glm2024chatglmfamilylargelanguage}; InternLM3-8B-Instruct~\citep{cai2024internlm2}; the Gemma3-it series (4B, 12B, 27B)~\citep{gemmateam2025gemma3technicalreport} and the Llama3.2 series (1B, 3B, 11B)~\citep{grattafiori2024llama3herdmodels}.

\subsection{Experimental Setup}
\paragraph{Inference configuration.} We run all 18 models with temperature 0.0 (greedy decoding) without \textit{thinking} and a single generation per instance. The GPT, DeepSeek and Kimi models are accessed through their respective official APIs while other models are deployed via a single H800. All prompts are written in Chinese; full prompt templates for each task are provided in Appendix~\ref{app:prompts}.

\paragraph{Scoring.} The Tier 2 category-specific tasks are scored deterministically: multiple-choice tasks are scored by option match, and open-ended restoration tasks by exact string match against the gold source form. Tier 1 (Definition Generation) is free-form and requires an automated judge. We use \texttt{Qwen3-235B-A22B-Instruct-2507}\footnote{\url{https://huggingface.co/Qwen/Qwen3-235B-A22B-Instruct-2507}} as the judge, prompted in Chinese to compare each model-generated definition against the reference definition and output a binary (0/1) correctness label. Empty or degenerate outputs are forced to 0 regardless of judge output. Judge reliability is validated against human annotators on a stratified sample of 1,000 instances, yielding 95\% agreement (Appendix~\ref{human_vs_judge}).

\subsection{Main Results}

\subsubsection{Tier 1: Definition Generation}

Table~\ref{definition_full} reports per-subcategory and overall definition generation accuracy for all 18 evaluated models. Three observations stand out.
\paragraph{Chinese neologisms remain an open challenge even for frontier models.} The top-performing model, Kimi-K2.5, reaches 67.74\% overall accuracy, and the majority of open-source models fall below 40\% overall accuracy. We observe a substantial fraction of \textsc{CNeo-Bench} entries, such as Homophonic and Lexical, remain systematically misinterpreted for models except APIs. This gap is not fully eliminated by scale alone within any single model family.

\paragraph{General frontier status does not transfer to Chinese 
neologisms.} GPT-5.1, despite its frontier status on general 
benchmarks, attains only 50.89\% overall on \textsc{CNeo-Bench}, below 
every Chinese-focused frontier model we evaluate. The gap is concentrated on subcategories that require Chinese-internal conventions, most visibly Pinyin Abbr. (GPT-5.1 at 47.40\%, Kimi-K2.5 at 76.56\%) and Unconv. Chars. (21.74\% vs.\ 82.61\%). This confirms that \textsc{CNeo-Bench} 
probes a capability tied to the specific conventions of Chinese 
neologisms, not captured by general frontier performance.

\paragraph{Subcategory difficulty is highly uneven.} 
Two patterns are particularly striking. First, English Abbr. are consistently easier than Pinyin Abbr. for most of the models (illustrated in Figure~\ref{enpinyin}, Appendix). This suggests that cross-script decoding (from Latin letters back to Chinese characters via Pinyin) poses a distinct challenge beyond general abbreviation understanding. Second, Unconv. Chars. shows the widest spread of any subcategory: most open-source models score below 20\%, yet Kimi-K2.5 reaches 82.61\%. Recovering a character from its visual components appears to be among the least uniformly acquired capabilities among different models.

\begin{table*}[t]
\centering
\small
\setlength{\tabcolsep}{5pt}
\renewcommand{\arraystretch}{1.15}
\resizebox{0.95\textwidth}{!}{
\begin{tabular}{l cc cc cc cc c c}
\toprule
 \multirow{2}{*}{\textbf{Model}} & \multicolumn{2}{c}{\textbf{Alphabetic}} & \multicolumn{2}{c}{\textbf{Homophonic}} & \multicolumn{2}{c}{\textbf{Lexical}} & \multicolumn{2}{c}{\textbf{Stylistics}} & \multirow{2}{*}{\textbf{Foreign}} & \multirow{2}{*}{\textbf{Avg.$^\ast$}} \\
\cmidrule(lr){2-3} \cmidrule(lr){4-5} \cmidrule(lr){6-7} \cmidrule(lr){8-9}
 & Pinyin & English & Chinese$^\dagger$ & Number$^\dagger$ & Lexical & Semantic & Unconv. Char.$^\dagger$ & Formulaic & \\
\midrule
GPT-5.1                 & 72.40 & \textbf{98.28} & 56.76 & 72.41 & 73.47 & 96.46 & 22.73 & 68.77 & \underline{86.36} & 75.70 \\
\addlinespace
DeepSeek-V3             & 77.60 & \textbf{98.28} & 65.25 & \underline{82.76} & 77.43 & \underline{96.58} & 45.45 & \underline{70.10} & 81.06 & 77.76 \\
DeepSeek-V3.2           & 76.04 & 94.83 & 64.46 & \underline{82.76} & 78.12 & 95.75 & \underline{54.55} & 64.20 & 81.31 & 75.61 \\
\addlinespace
Kimi-K2-Instruct        & \underline{78.65} & \underline{96.55} & \textbf{68.97} & \textbf{93.10} & \textbf{83.03} & \underline{96.58} & \textbf{59.09} & 68.17 & 84.60 & \underline{79.20} \\
Kimi-K2.5               & \textbf{80.21} & \underline{96.55} & \underline{68.17} & 79.31 & \underline{82.00} & \textbf{97.76} & \textbf{59.09} & \textbf{75.39} & \textbf{88.64} & \textbf{81.94} \\
\midrule
Qwen3-4B                & 23.44 & 86.21 & 9.28 & 0.00 & 49.53 & 86.79 & \underline{13.64} & 44.65  & 51.77 & 50.40 \\
Qwen3-4B-Instruct-2507  & 22.40 & 84.48 & 16.18 & 6.90 & 51.34 & 92.10 & \underline{13.64} & 46.03 & 54.04 & 52.99 \\
Qwen3-8B                & 30.21 & 86.21 & 19.10 & 3.45 & 52.63 & 90.21 & 4.55 & 49.22  &  58.59 & 54.97 \\
Qwen3-32B               & \underline{56.25} & \textbf{94.83} & \underline{33.69} & 13.79 & \underline{69.42} & \textbf{96.11} & \underline{13.64} & \underline{58.90} & \textbf{66.92} & \textbf{66.63} \\
\addlinespace
GLM-4-9B-0414           & 38.54 & 89.66 & 26.26 & 10.34 & 61.41 & 91.16 & 4.55 & 37.67 & 49.75 & 53.47 \\
GLM-4-32B-0414          & \textbf{65.62} & \underline{93.10} & \textbf{37.93} & \textbf{51.72} & \textbf{70.03} & 93.63 & \textbf{18.18} & 48.98 & \underline{66.67} & \underline{63.79} \\
\addlinespace
InternLM3-8B-Instruct   & 16.15 & 79.31 & 15.12 & 0.00 & 53.32 & 92.69 & 0.00 & \textbf{59.33} & 46.97 & 58.60 \\
\addlinespace
Gemma3-4B-it            & 30.73 & 86.21 & 9.55 & 3.45 & 39.79 & 91.04 & 4.55 & 29.60 & 44.95 & 43.22 \\
Gemma3-12B-it           & 39.58 & 89.66 & 21.22 & 10.34 & 58.66 & 92.69 & 4.55 & 44.65 & 57.07 & 55.78 \\
Gemma3-27B-it           & 41.67 & \textbf{94.83} & 29.44 & \underline{20.69} & 66.32 & \underline{95.64} & 4.55 & 48.13 & 62.37 & 60.71 \\
\addlinespace
Llama3.2-1B-Instruct    & 24.48 & 48.28 & 0.27 & 0.00 & 25.40 & 31.13 & 0.00 & 4.03  & 24.24 & 16.81 \\
Llama3.2-3B-Instruct    & 26.04 & 70.69 & 0.27 & 0.00 & 36.61 & 54.72 & 0.00 & 5.60  & 28.79 & 25.03 \\
Llama3.2-11B-Instruct   & 24.48 & 74.14 & 6.90 & 6.90 & 39.28 & 83.84 & 4.55 & 36.04  & 45.96 & 43.57 \\
\bottomrule
\end{tabular}}
\caption{Category-specific task results. 
Scores are reported as accuracy (\%). 
\textbf{Bold} indicates the best score in each column; \underline{underline} indicates the second best. 
Tasks with$^\dagger$~use open-ended restoration; all other tasks use multiple-choice format. Avg.$^\ast$ is computed as micro-average across all 4,759 instances.}
\label{tab:category-specific}
\end{table*}
\subsubsection{Tier 2: Category-Specific Diagnostics}
\label{sec:tier2}
Table~\ref{tab:category-specific} reports per-subcategory and overall Tier 2 accuracy for all 18 models. We highlight three observations.

\paragraph{Tier 2 accuracy is substantially higher than Tier 1 across the board.} The strongest model, Kimi-K2.5, reaches 81.94\% overall on Tier 2, and every model in our evaluation exceeds its own Tier 1 overall accuracy. This suggests that at least some of the neologism knowledge not surfaced in free-form description is nonetheless accessible under multiple choice framings or when contexts are given. However, for some small models, such as Qwen3-4B and Llama3.2-1B, their performance on tasks such as Pinyin Abbr. is statistically indistinguishable from random guessing.

\paragraph{Several MC subcategories approach saturation.} English Abbr. and Semantic Neo. are near-ceiling for most mid-to-frontier models. For English Abbr., this extends the pattern already visible in Tier 1, where English Abbr. are substantially easier than their Pinyin counterparts for models. For Semantic Neo., the task asks models to identify which of four sentences uses the target word in its novel sense, directly testing whether a model can distinguish a new sense from others. Accuracy on this subcategory is high across models, indicating that current LLMs can handle different senses in context well~\citep{ouyang2024llmsenseharnessingllmshighlevel}. Other MC subcategories (Formulaic., Foreign) show more gradual scaling, but still with room for improvement.

\paragraph{Open-ended subcategories show sharply different performance.} On Chinese Homo., Number Homo., and Unconv. Chars., 
frontier models achieve moderate accuracy while small-scale models 
frequently collapse to near-zero (e.g., InternLM3-8B scores 0\% on 
Number Homo. and Unconv. Chars.; Llama3.2-1B scores 0\% on Unconv. 
Chars.). With no random-guessing floor in the open-ended format, 
the frontier-to-small gap is widest on these tasks. We return to 
these subcategories in Section~\ref{sec:rm-gap} as their Tier 2 behavior reveals the most informative signal about what models can and cannot do beyond surface definition.

\section{Analysis}
To understand the patterns observed in former sections, we conduct two analyses: a contingency analysis between Tier 1 and Tier 2 outcomes (Section~\ref{sec:rm-gap}), and 
a few-shot mitigation experiment on hard samples (Section~\ref{sec:fewshot}).
We additionally provide an analysis 
on how humans perform on \textsc{CNeo-Bench} in 
Appendix~\ref{app:human-performance}.

\subsection{Recognition-Manipulation Gap}
\label{sec:rm-gap}
The per-subcategory comparison between Tier 2 and Tier 1 accuracy 
(Figure~\ref{fig:heatmap}, Appendix) shows that most subcategories 
have uniformly positive gaps across models, reflecting the 
task-format advantages discussed in Section~\ref{sec:tier2}. 
Three subcategories diverge from this pattern: Chinese Homo., 
Number Homo., and Unconv. Chars., precisely the open-ended 
restoration subcategories. We focus on these three below.
\begin{figure*}[t]
  \centering
  \includegraphics[width=0.99\textwidth]{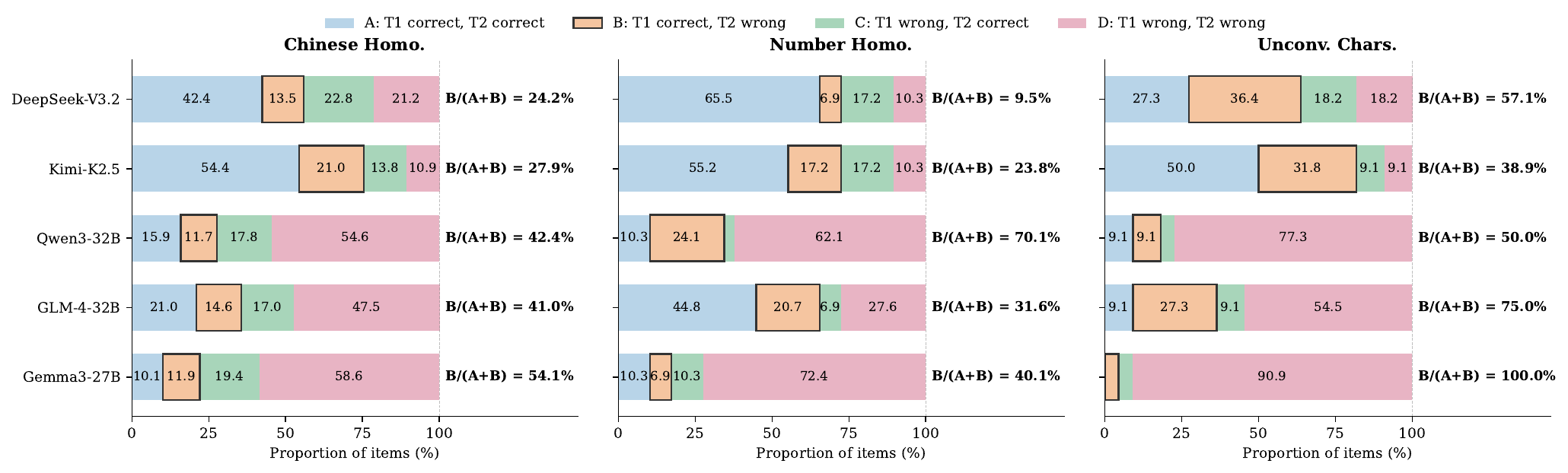}
  \caption{Contingency analysis of Tier 1 (definition generation) vs Tier 2 (category-specific) outcomes for five representative models on three open-ended subcategories. Cells A/B/C/D denote (T1$\checkmark$, T2$\checkmark$), (T1$\checkmark$, T2$\times$), (T1$\times$, T2$\checkmark$), (T1$\times$, T2$\times$) respectively. The highlighted B/(A+B) column reports the operation failure rate (the fraction of correctly described neologisms on which restoration fails) directly quantifying the recognition-manipulation gap.}
  \label{figure_cell}

\end{figure*}

We refer to this systematic divergence: in which models can describe a neologism in Tier 1 yet fail to produce its source form in Tier 2, as the \textbf{recognition-manipulation gap (RM Gap)}. To quantify it, we conduct 
a per-item contingency analysis between Tier 1 and Tier 2 outcomes, 
partitioning items into four cells: \textbf{A} (T1 $\checkmark$, 
T2 $\checkmark$), \textbf{B} (T1 $\checkmark$, T2 $\times$), \textbf{C} (T1 $\times$, 
T2 $\checkmark$), and \textbf{D} (T1 $\times$, T2 $\times$). The fraction 
$B/(A+B)$ quantifies the RM Gap as an \textit{operation failure rate}: the proportion of items the 
model can correctly describe in Tier 1 but fails to restore in 
Tier 2. Figure~\ref{figure_cell} reports the cell distribution 
for five representative models.

\paragraph{Operation failure exists across models and subcategories.} Figure~\ref{figure_cell} 
shows substantial operation failure across five representative models and 
three subcategories. On Chinese Homo., despite shared writing 
system and an explicit contextual example, $B/(A+B)$ ranges from 
\textbf{24.2\%} (DeepSeek-V3.2) to \textbf{54.1\%} (Gemma3-27B), even 
Kimi-K2.5, the strongest Tier 1 model, fails to restore 
\textbf{27.9\%} of items it can describe. Number Homo. shows similar failure rates across models. Unconv. Chars. shows 
the most pronounced failure: \textbf{38.9\%} for Kimi-K2.5, 
\textbf{57.1\%} for DeepSeek-V3.2, and over \textbf{50\%} for other models.  Notably, high failure rates occur on the models with strong Tier 1 accuracy.

\paragraph{Inside Cell B: how restoration fails.}
To understand what the RM Gap looks like 
in practice, we inspect the Cell B records and classify each 
failure into one of four modes: \textbf{copy verbatim} (model 
outputs the neologism itself), \textbf{explanation/paraphrase} 
(model outputs a semantic gloss or synonym of the source form), 
\textbf{wrong restoration} (output semantically unrelated to the 
source form), and \textbf{empty/other}. The full breakdown is 
reported in Table~\ref{tab:failure-mode-breakdown}, Appendix. Across all 
three subcategories, the dominant failure mode is 
\textbf{explanation or paraphrase}: rather than producing the 
source form, the model outputs a semantic equivalent of it. 
Table~\ref{tab:case-study} shows representative cases of this 
failure mode. For example, Kimi-K2.5 correctly recognizes ``石乐志'' as a homophone of ``失了智'' (lose one's wits) in Tier 1, yet in Tier 2 outputs ``失智'' (dementia) rather than the source form. 

\paragraph{Cell C reflects in-context inference.} Cell C stands for items 
the model fails to describe in Tier 1 but answers correctly in 
Tier 2. On Chinese Homo. and Number Homo., where Tier 2 
provides a usage context, inspection of Cell C records suggests 
that many successes reflect in-context inference of meaning from 
the example sentence. Tier 2 accuracy on context-augmented 
subcategories should therefore be interpreted as performance 
with additional contextual support, not as evidence of memorized 
neologism knowledge.

% \paragraph{Implications.} These three findings depend on the mechanism-driven taxonomy of \textsc{CNeoBench}. An evaluation that aggregates open-ended and MC subcategories together would produce a uniformly positive Tier 2 gap and miss all three patterns: the MC subcategories' positive gaps would dominate any average, the scale-dependent flip on Number Homo.\ would average toward zero, and the concentrated frontier failures on Unconv. Chars.\ would be diluted by the near-zero gaps of smaller models. The taxonomy is what makes these failures visible. The findings also suggest a direction for benchmark extension: richer diagnostic tasks for Chinese neologisms should include more tasks that require producing specific source forms rather than recognizing them from options, especially for mechanisms that involve crossing representational substrates (phonetic, visual, cross-symbol). We return to this in Section~\ref{sec:discussion}.
\begin{figure*}[t]
  \centering
  \includegraphics[width=0.99\textwidth]{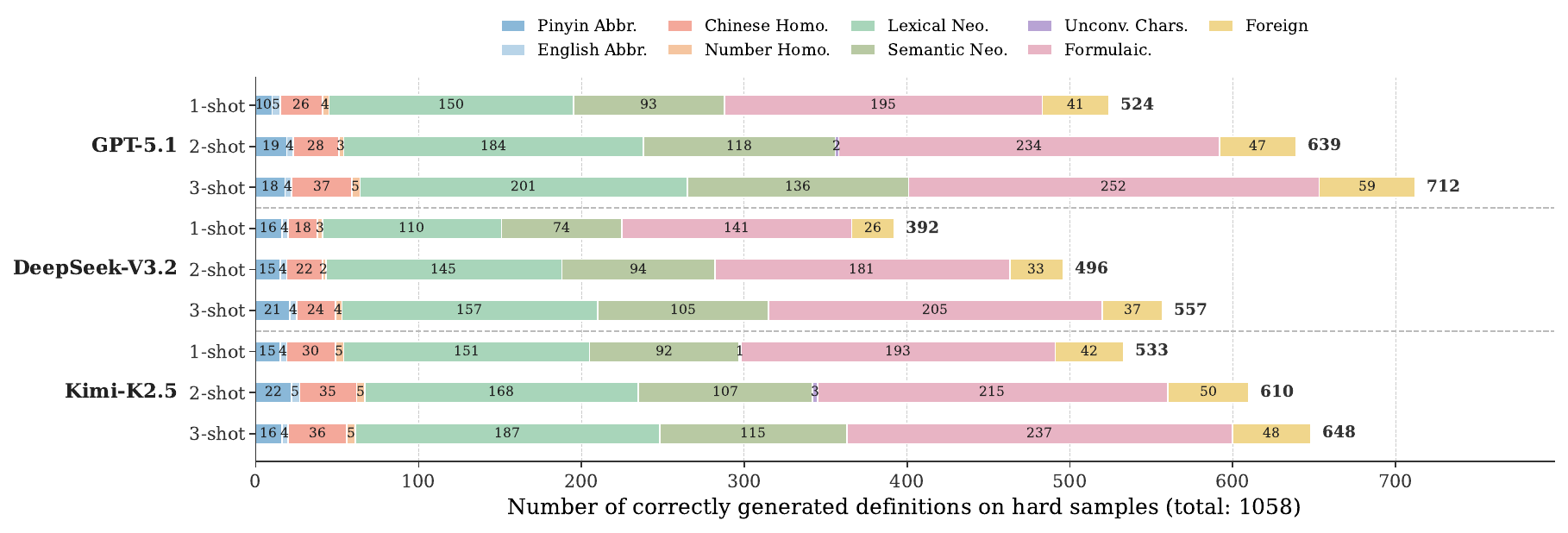}
  \caption{Few-shot recovery on 1{,}058 hard samples. Each bar shows the number of hard samples recovered (i.e., judged correct under the few-shot setting) under 1-shot, 2-shot, and 3-shot prompting. Bars are stacked by subcategory to show the distribution of recoveries across the nine subcategories.}
  \label{fewshot_understand}

\end{figure*}
\paragraph{Implications from RM Gap.} Two mechanisms plausibly contribute to 
the recognition-manipulation gap, and the observed failure mode 
is consistent with both. \textbf{First}, neologisms appear in pretraining 
data predominantly as standalone lexical items, which are used naturally 
in a complete sentence, often without an accompanying explanation 
that links them back to their source form. A model trained on this 
distribution can readily produce a description of what a neologism 
means, yet have no reliable retrieval path from the neologism to 
the specific surface form it was derived from. \textbf{Second}, the task 
of source-form \textit{restoration} is itself rare in pretraining: 
the model may encounter contexts of what a neologism means, but does not truly understand the relationship between the source form and the neologism. When restoration is explicitly prompted, the model defaults to its dominant trained behavior: producing meaning rather than the form the task demands. These two mechanisms reinforce each other: the data does not equip models with the source-form mapping, and restoration is not an operation models are trained to perform even when it is explicitly required.

\subsection{Hard Samples and Few-Shot Recovery}
\label{sec:fewshot}
\paragraph{Hard samples.} We define hard samples as entries on which GPT-5.1, DeepSeek-V3.2, and Kimi-K2.5 all fail the Tier 1 definition generation task under zero-shot prompting. This yields 1,058 entries, 22.23\% of the benchmark. The hard rate is roughly balanced across subcategories (Table~\ref{statistics}), indicating that the remaining difficulty is distributed across multiple formation mechanisms rather than concentrated in any single type. 

% Figure~\ref{fig:venn} shows the error overlap among the three models; pairwise Jaccard similarities (0.45--0.60) confirm that the three models fail on substantially overlapping but non-identical subsets, and that the 1,058 joint-failure set is not an artifact of a single model's weakness.
\vspace{-2mm}
\paragraph{Few-shot recovery.} We investigate whether the difficulty of hard samples reflects absence of in-context exposure or absence of underlying capability. For each hard sample, we construct 1 to 3 natural usage sentences (generated by an independent LLM given the neologism and its reference definition, then manually filtered), and rerun Tier 1 definition generation with these sentences as few-shot exemplars. Figure~\ref{fewshot_understand} reports the number of recovered hard samples across three shot counts on the same three frontier models. Even 1-shot is sufficient to recover 37--50\% of hard samples across models; 3-shot recovers 53--67\%. 
% Marginal gains decrease with additional shots (e.g., GPT-5.1: +115 from 1$\to$2-shot, +73 from 2 $\to$3-shot), indicating that a small number of in-context uses saturates most of the recoverable signal.
\vspace{-2mm}
\paragraph{Interpretation.} Two points are worth noting. \textbf{First}, the substantial recovery at 1-shot suggests that failure on most hard samples reflects insufficient pretraining exposure to specific neologisms rather than a fundamental capability gap: the knowledge needed is present in the model but not readily activated without contextual grounding. \textbf{Second}, even with 3-shot context, 33--47\% of hard samples remain unrecovered across all three models. These residual failures suggest that scaling exposure alone is insufficient: progress on Chinese neologisms involving phonetic or visual mechanisms may require training approaches that explicitly represent these substrates, rather than relying on surface-form generalization.

\section{Related Work}
Early work of neologisms mainly focuses on identification in English text 
\citep{cook-stevenson-2010-automatically-identifying, mccrae-2019-identification-neologisms, zalmout-etal-2019-unsupervised-neologism-norm}. 
With the rise of LLMs, recent benchmarks evaluate whether models 
can handle these neologisms
\citep{zheng-etal-2024-neobench}. Existing efforts focus largely on 
English neologisms, and do not engage with 
mechanisms specific to the linguistic features behind the production of neologisms. 
Chinese neologism evaluation has emerged primarily in the context 
of machine translation, where new expressions pose translation 
challenges due to their cultural specificity and rapid evolution 
\citep{qian-etal-2024-evaluating, guo2025redefiningmachinetranslationsocial, miao2026neoamt, zhao2026benchmarkingmachinetranslationchinese}. 
A small body of work examines whether LLMs understand Chinese 
neologisms directly \citep{CanLargeLanguageModelsUnderstandChineseNeologisms}, but is relatively limited in evaluation scope without 
an explicit taxonomy of formation mechanisms. \textsc{CNeo-Bench} 
fills this gap by organizing the evaluation around the 
mechanisms, and extends evaluation in scale and granularity.

% \input{tables/definition_generation}

% Bibliography entries for the entire Anthology, followed by custom entries
%\bibliography{anthology,custom}
% Custom bibliography entries only

\section{Conclusion}
We introduce \textsc{CNeo-Bench}, a benchmark of 4,759 Chinese neologisms with reference definitions, organized by the linguistic mechanism and designed in a two-tier evaluation framework. Evaluating 18 LLMs, we find that Chinese neologism understanding remains an open challenge, and that few-shot prompting cannot fully resolve it. In addition, we identify a systematic recognition-manipulation gap (describe correctly in Tier 1 but fail in Tier 2) that surfaces across models. \textsc{CNeo-Bench} thus provides a mechanism-aware lens for evaluating Chinese-language LLMs and lays the groundwork for future benchmarks that go beyond describing meaning alone.

\clearpage

\section*{Limitations}
Our work has several limitations worth noting. First, Tier 1 
correctness is judged by an LLM judge on semantic adequacy. Although has been compared with human evaluation, a definition may be 
marked correct if it accurately conveys the meaning of the 
neologism, without explicitly reproducing the source form of it. 
Second, two subcategories: Number Homo. ($n = 29$) and Unconv. 
Chars. ($n = 23$), contain few items because the underlying 
neologism types are themselves rare in Chinese neologisms and our collection covered the 
items we could verify. Fine-grained model-by-model comparisons on 
these two subcategories should therefore be interpreted as 
indicative rather than definitive. Third, our work focuses on diagnosing the recognition-manipulation gap through evaluation, and we leave mitigation strategies to future work.

\section*{Ethical Statements}
\textsc{CNeo-Bench} is constructed from publicly available sources, 
including Moegirlpedia, Wikiversity, and other open web content 
indexed by general-purpose search. We verified the licensing of 
each source and included only content that permits research use 
with attribution. The benchmark contains no personally identifiable 
information, and neologisms involving real public figures are 
limited to the public-domain context in which the expressions 
themselves originated. We use LLMs only for grammar 
and language polishing during manuscript preparation; all 
substantive content was written by the authors.

\bibliography{custom}

\appendix

\section{Appendix}
\label{sec:appendix}

\subsection{Detailed Taxonomy}
\label{detailed_tax}
A summary of examples and explanations for each subcategory is provided in Table~\ref{category_examples}.

\paragraph{(1) Alphabetic Abbreviations.} Expressions formed by abbreviating a full phrase into a sequence of initial letters. We distinguish \textit{Pinyin Abbr.}, where the initials are drawn from the Pinyin transliteration of a Chinese phrase (\textit{yyds} ← 永远的神), from \textit{English Abbr.}, where the initials are drawn from an English phrase already in circulation among Chinese speakers (\textit{ACG} ← \textit{Anime, Comics, Games}). Although the two subcategories share a surface form (a short Latin-letter string), they differ sharply in the inferential chain required to decode them, motivating their separation.

\paragraph{(2) Homophonic Expressions.} Expressions that encode a target phrase through phonetic substitution. \textit{Number Homo.} uses digit sequences whose spoken form approximates the target Chinese phrase (886 ← 拜拜咯, bye bye), exploiting the phonetic mapping between Mandarin number words and common colloquial expressions. \textit{Chinese Homo.} replaces characters of the target phrase with different characters that share a similar pronunciation (雨女无瓜 ← 与你无关, none of your business), producing a written form that is orthographically unrelated to but phonetically reminiscent of the source.

\paragraph{(3) Lexical Innovations.} Expressions whose novelty lies at the lexical-semantic level. \textit{Lexical Neo.} introduces newly coined items that carry a previously unattested meaning (键盘侠, keyboard warrior). \textit{Semantic Neo.} retains an existing word form but shifts it to a new meaning in contemporary usage (养鱼, literally fish farming, now used to describe maintaining multiple romantic prospects simultaneously). The two subcategories differ in whether the innovation operates on form or on sense. Lexical Neo. creates expressions with no prior meaning in standard Chinese, whereas Semantic Neo. reuses existing words but conveys a new meaning.

\paragraph{(4) Stylistic and Pragmatic Expressions.} Expressions whose form is conventionally fixed rather than generated by a productive rule. \textit{Unconv. Chars.} decomposes a standard Chinese character into its visual components and writes the components separately as a substitute for the original character (彳亍 ← 行, okay: the character 行~is split into its left part 彳 and right part 亍). The decomposition is a community-accepted convention for specific characters rather than a rule that can be freely applied to any character. \textit{Formulaic} expressions are full phrases or sentences, typically originating from a viral video, post, or meme, that are reused as a single unit to convey a specific pragmatic effect (没活了可以咬打火机, used to mock someone who has run out of content to produce). What these two share is that their meaning cannot be worked out compositionally from the parts: a reader either recognizes the specific form or does not, with no rule-based fallback.

\paragraph{(5) Foreign Borrowings.} Expressions derived from foreign-language sources. This category encompasses several generation paths, such as phonetic reinterpretation, where a foreign phrase is rewritten using Chinese characters chosen for their phonetic approximation to the source (逮虾户 ← \textit{Déjà Vu}, via the Japanese anime \textit{Initial D}), and direct lexical adoption, typically from Japanese (文化祭, cultural festival, borrowed directly from Japanese characters). These paths share the same defining feature of this category: dependence on a non-Chinese source, which motivates treating them as a single subcategory.

\subsection{Annotation}
\label{annotation}
The three annotators are all university-aged native Mandarin speakers from mainland China who familiar with contemporary Chinese internet culture. Annotators are trained by the authors using a written rubric derived from the taxonomy definitions. Raw three-way agreement on the 9-category scheme is 78\% (Fleiss' $\kappa$=0.69). For the remaining 22\% with at least one disagreement, a senior reviewer assigned the final label through adjudication following the same rubric. The same reviewer then performed a quality-control pass over the entire dataset, identifying and correcting an additional 4\% of labels, to yield the final gold label set. All annotators are paid with standard hourly rate. 

The annotation instructions are originally delivered in Chinese and are 
presented below in both the original Chinese and an English 
translation.
\begin{promptbox}{Annotation Task}
\textbf{Chinese (original):}

\vspace{2pt}
任务说明:你将看到一个中文网络新词以及它对应的参考释义。你的任务是根据该新词的构成机制,从下列 9 个子类中选出最合适的一个,作为该词的子类标签。如果你认为某个新词不能确定归属(可能同时符合两个子类的特征,或不符合任何子类),请打上 "uncertain" 标签,并简要说明你的疑问,后续由资深标注者(senior annotator)审核确认。

\vspace{6pt}
\textbf{English (translation):}

\vspace{2pt}
\textit{Task description: You will see a Chinese internet 
neologism and its corresponding reference definition. Your 
task is to choose one of the nine subcategories below as the 
subcategory label for that neologism, based on the mechanism 
behind its formation. If you cannot confidently assign a 
neologism to a single subcategory (e.g., it appears to match 
multiple subcategories, or does not clearly fit any), mark it 
as ``uncertain'' with a brief note on your reasoning; a senior 
annotator will review and finalize the label.}
\end{promptbox}

Annotators are provided with Table~\ref{category_examples} as the definition for nine subcategories.

\subsection{Human Evaluation of LLM-as-Judge}
\label{human_vs_judge}

To validate the use of an LLM judge for scoring Tier 1 definition 
generation outputs, we conduct a human evaluation on a stratified 
sample of model outputs.

\paragraph{Setup.} We sampled 1,000 model outputs from 
the Tier 1 generation results, stratified across the nine 
subcategories to ensure coverage of all formation mechanisms. 
Subcategories with small total sizes such as Unconv. Chars. are exhaustively included in the evaluation set. The model outputs 
covered all 18 models. 

\paragraph{Human Annotation.} The same three annotators who labeled subcategories in Appendix~\ref{annotation} independently judged whether each model-generated definition correctly described the meaning and origin of the target neologism. The judging criterion mirrored the criterion given to 
the LLM judge (Appendix~\ref{app:judge-prompt}). The final human label for each item was obtained by majority vote.

\paragraph{Agreement Results.} The LLM judge agreed with the 
majority-voted human label on \textbf{95\%} of the 1,000 validation 
items (Cohen's $\kappa$ = 0.74). The remaining 5\% of 
disagreement cases primarily involve borderline definitions where 
the boundary between `correct' and `incorrect' is itself ambiguous, 
such as definitions that convey approximate meaning but miss 
specific source-form references. We also note that a small number of items where the model produced an empty output were incorrectly judged as correct by the LLM judge. These cases are rare but contribute to the lenient bias discussed above.
\begin{figure}[ht]
  \centering
  \includegraphics[width=0.49\textwidth]{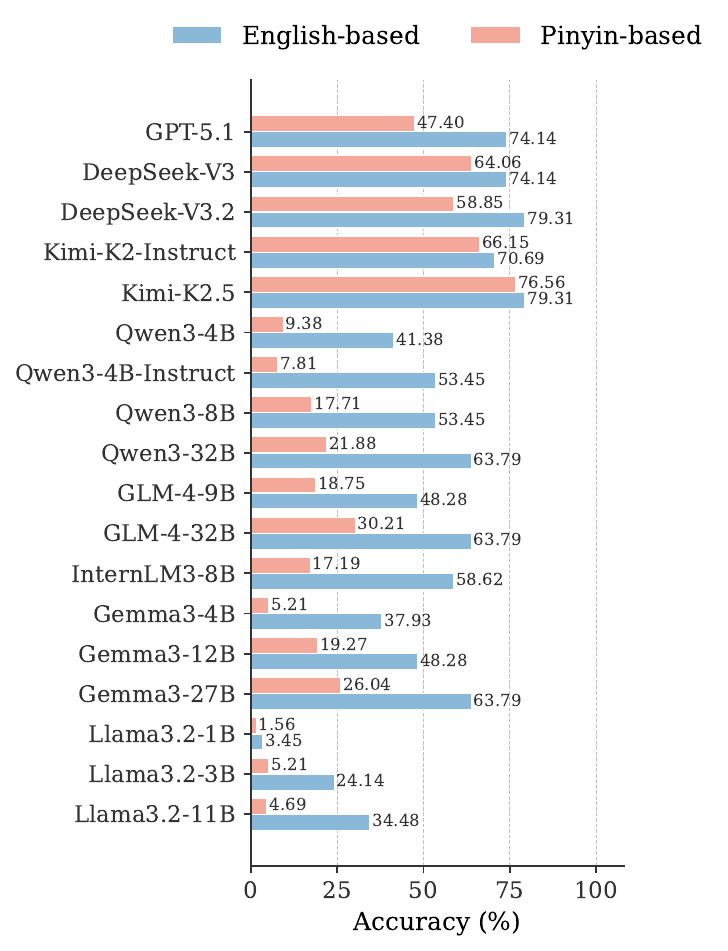}
  \caption{Accuracy on English-based vs Pinyin-based alphabetic abbreviations in the definition generation task. English-based abbreviations are substantially easier than Pinyin-based ones for all evaluated models.}
  \label{enpinyin}

\end{figure}

\subsection{Human Performance on \textsc{Cneo-Bench}}
\label{app:human-performance}
To calibrate the difficulty of \textsc{CNeo-Bench} and provide a 
human reference for Tier 1 definition generation, we conducted a 
sample-based human evaluation.

\paragraph{Setup.} 
Three native Mandarin speakers familiar with Chinese 
internet neologisms participated. For each of the nine subcategories, 
we randomly sampled 100 items when the subcategory contained more 
than 100 items; subcategories with fewer than 100 items were evaluated in 
full. 

\paragraph{Evaluation Procedure.} 
Evaluators were presented with each neologism in isolation, 
without access to its reference definition, and asked to write 
a definition in their own words, mirroring the Tier 1 generation 
task setup. The human-written definitions were scored 
using the same LLM judge and the same prompt as in our main experiment. Each item was evaluated by three evaluators, with the final score taken as the average of the judged scores.

\paragraph{Results.} 
Human evaluators achieved an overall accuracy of 85.63\%
on Tier 1, compared with 67.74\% for the strongest LLM 
(Kimi-K2.5). Per-subcategory breakdowns are reported in 
Table~\ref{tab:human-performance}. 

\begin{table}[t]
\centering
\small
\setlength{\tabcolsep}{6pt}
\renewcommand{\arraystretch}{1.2}
\begin{tabular}{lcc}
\toprule
\textbf{Subcategory}  & \textbf{Human} & \textbf{Kimi-K2.5} \\
\midrule
Pinyin Abbr.         & 91.00\% & 76.56\% \\
English Abbr.        & 84.48\% & 79.31\% \\
Chinese Homo.        & 94.00\% & 70.82\% \\
Number Homo.         & 82.76\% & 72.41\% \\
Lexical Neo.         & 87.00\% & 66.06\% \\
Semantic Neo.        & 83.00\% & 67.06\% \\
Unconv. Chars.       & 95.65\% & 82.61\% \\
Formulaic            & 78.00\% & 67.55\% \\
Foreign              & 80.00\% & 61.69\% \\
\midrule
\textbf{Overall}     & \textbf{85.63\%} & \textbf{67.74\%} \\
\bottomrule
\end{tabular}
\caption{Human performance on a stratified sample of 
\textsc{CNeo-Bench} Tier 1 definition generation. We leave Kimi-K2.5, the strongest model we evaluate, for comparison.}
\label{tab:human-performance}
\end{table}

Inspection of the human evaluation errors reveals two recurring 
failure patterns. The first is \textbf{recognition failure}: 
since Tier 1 presents only the neologism without contextual usage, 
evaluators occasionally encounter terms they have not previously 
seen and are unable to produce a correct definition. This pattern 
is most visible on Formulaic., Lexical Neo., and Semantic Neo. 
subcategories, where some expressions may be outside any given 
evaluator's exposure. 

The second pattern is specific to \textbf{Foreign Borrowings} and 
reflects a stricter judging criterion on this subcategory. Several 
human-written definitions correctly convey the meaning of a 
borrowed expression but omit the specific source language or 
borrowing path (e.g., glossing ``哈基米'' as ``cat'' without 
identifying its origin in Japanese), causing the 
LLM judge to mark them incorrect. 
\begin{figure*}[t]
  \centering
  \includegraphics[width=0.99\textwidth]{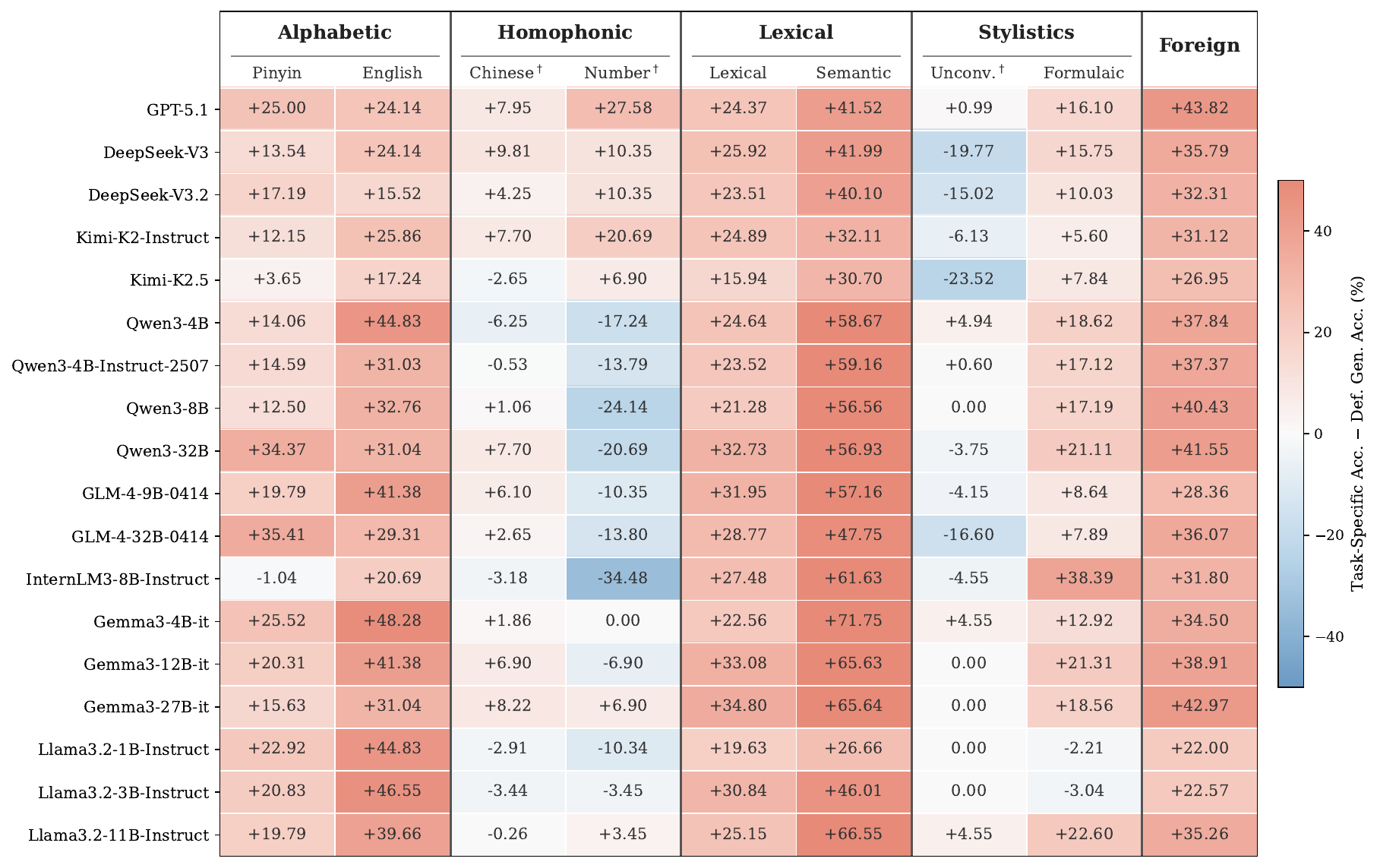}
  \caption{Tier 2 minus Tier 1 accuracy (percentage points) across 18 models and 9 subcategories. Positive gaps (red) indicate Tier 2 exceeds Tier 1; negative gaps (blue) indicate Tier 2 underperforms Tier 1. Subcategories marked with
$^\dagger$ use open-ended restoration; others use multiple-choice.}
  \label{fig:heatmap}

\end{figure*}

\renewcommand\tabularxcolumn[1]{m{#1}}
\begin{table*}[htbp]
\centering

\small
\begin{tabularx}{\textwidth}{>{\centering\arraybackslash}m{2.2cm} 
                              >{\centering\arraybackslash}m{2.2cm} 
                              >{\raggedright\arraybackslash}X 
                              >{\raggedright\arraybackslash}X}
\toprule
\textbf{Categories} & \textbf{Subcategories} & \textbf{Explanations} & \textbf{Examples} \\
\midrule

\multirow{3}{=}[-1.2em]{1.~Alphabetic Abbreviations} 
& Pinyin Abbr. 
& Expressions formed by taking the initial letters of the Pinyin transliteration of Chinese phrases to create abbreviated forms. 
& \textit{yyds} from ``永远的神'' (yong yuan de shen), which means forever a legend. \\
\cmidrule(l){2-4}
& English Abbr.
& Expressions formed by taking the initial letters of the English phrases to create abbreviated forms. 
& \textit{ACG} from ``Anime, Comics, Games''. \\
\midrule

\multirow{2}{=}[-1.1em]{2.~Homophonic Expressions} 
& Number Homo.
& Expressions that use numbers to convey meaning based on phonetic similarity. 
& ``拜拜咯'' becomes ``886'' as they have similar pronunciations in Chinese, which means bye bye. \\
\cmidrule(l){2-4}
& Chinese Homo.
& Expressions that replace standard Chinese characters with characters of similar pronunciation. 
& ``与你无关'' (yu ni wu guan) becomes ``雨女无瓜'' (yu nv wu gua), which means none of your business. \\
\midrule

\multirow{2}{=}[-1.9em]{3.~Lexical Innovations} 
& Lexical Neo.
& Newly coined expressions that introduce novel lexical items. 
& ``键盘侠'', which means keyboard warrior, a person who aggressively criticizes or attacks others online without engaging in real-world action. \\
\cmidrule(l){2-4}
& Semantic Neo.
& Existing words that acquire new meanings in emerging usage. 
& ``养鱼'', literal meaning: fish farming; new meaning: to maintain multiple romantic prospects simultaneously, often by keeping others as backups. \\
\midrule

\multirow{2}{=}[-1.8em]{4.~Stylistic and Pragmatic Expressions} 
& Unconv. Chars. 
& Expressions that employ rare, archaic, or visually stylized Chinese characters in non-standard ways to convey meaning or achieve stylistic and expressive effects. 
& ``彳亍'', an expression composed of two archaic characters ``彳'' and ``亍'' to form ``行'', which means okay. \\
\cmidrule(l){2-4}
& Formulaic
& Fixed or semi-fixed expressions derived from specific memes or events, which are reused across contexts to convey particular pragmatic meanings. 
& ``没活了可以咬打火机'', a meme-based expression originating from a viral livestream clip, used to humorously suggest that someone has run out of ideas or content. \\
\midrule

% \multirow{2}{=}[-1.2em]{5.~Foreign Language Borrowings} 
% & Homophonic 
% & Foreign phrases that are reinterpreted as Chinese characters based on phonetic similarity. 
% & ``逮虾户'', a phonetic expression from ``Deja Vu'' (originates from the Japanese Anime, \textit{Initial D}), is used to depict high-speed driving or drifting scenes. \\
% \cmidrule(l){2-4}
% & Loanwords 
% & Expressions derived from foreign languages through adoption or extension. 
% & ``文化祭'', a loanword from Japanese referring to a school cultural festival. \\

\multirow{1}{=}[0.5em]{5.~Foreign Borrowings}
& Foreign$^{\ast}$ 
& Expressions derived from foreign-language sources through phonetic reinterpretation or direct lexical adoptiont. 
& ``逮虾户'' from ``Déjà Vu'' (phonetic, via Japanese anime \textit{Initial D}), used for high-speed driving scenes; ``文化祭'', which means school cultural festival is a direct loanword from Japanese. \\

\bottomrule
\end{tabularx}
\caption{Mechanism-driven taxonomy of Chinese neologisms in \textsc{CNeo-Bench}, with five top-level categories and nine subcategories. $^{\ast}$Foreign Borrowings has no further subcategorization and is treated as a single class.}
\label{category_examples}
\end{table*}

\begin{table*}[t]
\centering
\small
\setlength{\tabcolsep}{6pt}
\renewcommand{\arraystretch}{1.3}
\newcolumntype{T}{>{\hsize=0.4\hsize}X}  % Task narrow
\newcolumntype{W}{>{\hsize=2.3\hsize}X}    % Example wide
\newcolumntype{N}{>{\hsize=0.3\hsize}X}  % Answer narrow
\resizebox{0.9\textwidth}{!}{
\begin{tabularx}{\textwidth}{l T W N}
\toprule
\textbf{Subcategory} & \textbf{Task} & \textbf{Example} & \textbf{Answer} \\
\midrule

Pinyin Abbr. & Cloze & 
\textit{Context}: 这家楼下新开的烧烤摊味道真的太绝了，朋友吃完直接说它是夜宵界的(\_\_\_\_)。 \newline 
\textit{English}: The new BBQ stall that just opened downstairs is absolutely amazing, my friend straight-up called it the (\_\_\_\_) of the late-night food scene. \newline
\textit{Options}: (A) cxk  (B) xswl  (C) yysy  (D) yyds & 
(D) yyds \newline(``forever legendary'') \\
% \addlinespace
\midrule
English Abbr. & Cloze & 
\textit{Context}: 他平时看番、追漫画还爱玩手游，朋友圈里一看就是个(\_\_\_\_)重度爱好者。 \newline
\textit{English}: He watches anime, follows manga, and loves playing mobile games, and you can tell he's a hardcore (\_\_\_\_) enthusiast from his social media. \newline
\textit{Options}: (A) KPI  (B) ACG  (C) PDF (D) CPU & 
(B) ACG \newline (Anime, Comics, Games) \\
% \addlinespace
\midrule
Chinese Homo.$^\dagger$ & Restoration & 
\textit{Neologism}: 雨女无瓜 \newline
\textit{Example context}: 这件事是他们部门内部的安排，雨女无瓜，别再到处打听了。\newline
\textit{English}: This is an internal arrangement within their department, none of your business, so stop asking around about it.
& 
与你无关 \\
%\addlinespace
\midrule
Number Homo.$^\dagger$ & Restoration & 
\textit{Neologism}: 886 \newline
\textit{Example context}: "时间不早了, 我先撤, 886, 明天见。 \newline
\textit{English}: It's getting late, I'm gonna head out. Bye-bye, see you tomorrow!
& 
拜拜咯 \\
%\addlinespace
\midrule
Lexical Neo. & Cloze & 
\textit{Context}: 事情刚上热搜、真相还没出来，他就在评论区里对当事人冷嘲热讽、指点江山，活脱脱一个(\_\_\_\_)。 \newline
\textit{English}: The story had just started trending and the truth hadn't even come out yet, but there he was in the comments section mocking the people involved and pontificating about everything, a textbook (\_\_\_\_). \newline
\textit{Options}: (A) 程序员   (B) 黑客  (C) 杠精  (D) 键盘侠 & 
(D) 键盘侠 \newline (keyboard warrior) \\
% \addlinespace
\midrule
Semantic Neo. & Semantic \newline Discrimination & 
\textit{Target word}: 养鱼 \newline
\textit{Sentences}: \newline
(A) 他在院子里挖了个小池子，专门养鱼和种睡莲。(fish farming) \newline
(B) 爷爷退休后最大的乐趣就是养鱼，每天一早都去喂食。(fish farming) \newline
(C) 她总是同时和好几个人保持暧昧，朋友都说她是在养鱼。 \newline
(D) 这家农场除了养鸡养鸭，还养鱼，逢年过节卖得特别好。(fish farming) & 
(C) maintaining multiple romantic relationships \\
% \addlinespace
\midrule
Unconv. Chars.$^\dagger$ & Restoration & 
\textit{Neologism}: 彳亍 & 
行 (``okay'') \\
% \addlinespace
\midrule
Formulaic & Scenario \newline Matching & 
\textit{Scenario}: 家庭聚会上，62岁的老周本来在陪孙子搭积木，结果看到院里积雪够厚，突然兴奋地拉着几个晚辈去堆``巨型雪怪''，还非要拿胡萝卜做獠牙。被老伴笑着数落一把年纪还这么能闹，他却越玩越起劲。 \newline
\textit{English}: At the family gathering, 62-year-old Zhou had been building blocks with his grandson, but when he saw how thick the snow had piled up in the yard, he suddenly got excited and dragged a few of the younger family members out to build a ``giant snow monster'', and he insisted on using carrots for the fangs. His wife laughingly scolded him for being so rowdy at his age, but the more he played, the more into it he got. \newline
\textit{Options}: (A) 老夫聊发少年狂  (B) 人间清醒  (C) 差生文具多  (D) 男人至死是少年 & 
(D) 男人至死是少年 \newline (Men never really grow up) \\
% \addlinespace
\midrule
Foreign$^\dagger$ & Source \newline Identification & 
\textit{Neologism}: 文化祭 \newline
\textit{Options}: \newline
(A) 它是从日语``文化祭(ぶんかさい)''转译扩展的词，``祭''在日语里有集中展示、主题活动之意，``文化祭''本质上是对日本校园文化展示周这一活动与说法的概括吸收。(Expanded translation)\newline 
(B) 它是从日语``文化祭(ぶんかさい)''直接借入的词，``祭''在日语里有节庆、庆典之意，``文化祭''本质上是对日本校园``学园文化祭''这一制度与说法的直接沿用。(Direct loanword)  \newline
(C) 它是从日语``文化祭(ぶんかさい)''类推形成的词，``祭''在日语里有公开活动、校内集会之意，``文化祭''本质上是对日本校园``社团发表会''这一形式与说法的重新命名。(Analogical formation) \newline
(D) 它是从日语``文化祭(ぶんかさい)''借形改义的词，``祭''在日语里有热闹活动、集体庆祝之意，``文化祭''本质上是对日本校园``校园开放日''这一安排与说法的本地化套用。(Borrowed form, new meaning)
& 
(B) Direct loanword \\

\bottomrule
\end{tabularx}}
\caption{Representative examples for each Tier 2 tasks in \textsc{CNeo-Bench}. Subcategories marked with $^\dagger$ require the model to complete open-ended restoration. \textit{English} references are not included in our dataset.}
\label{tab:task-examples}
\end{table*}

\begin{table*}[t]
\centering
\small
\setlength{\tabcolsep}{6pt}
\renewcommand{\arraystretch}{1.3}
\begin{tabular}{ll cccc}
\toprule
\multirow{2}{*}{\textbf{Model}} & \multirow{2}{*}{\textbf{Subcategory}} 
& \multicolumn{4}{c}{\textbf{Cell B Failure Mode (\%)}} \\
\cmidrule(lr){3-6}
& & Copy Verbatim & Explanation/Paraphrase & Wrong Restoration & Empty/Other \\
\midrule
\multirow{3}{*}{Kimi-K2.5}
  & Chinese Homo.    & 26.6\% & \textbf{38.0\%} & 32.9\% & 2.5\% \\
  & Number Homo.     & 0.0\% & \textbf{60.0\%} & 40.0\% & 0.0\% \\
  & Unconv. Chars.   & 0.0\% & \textbf{71.4\%} & 28.6\% & 0.0\% \\
\addlinespace
\multirow{3}{*}{DeepSeek-V3.2}
  & Chinese Homo.    & 25.5\% & \textbf{35.3\%} & 35.3\% & 3.9\% \\
  & Number Homo.     & 0.0\% & \textbf{100\%} & 0.0\% & 0.0\% \\
  & Unconv. Chars.   & 0.0\% & \textbf{62.5\%} & 37.5\% & 0.0\% \\
\addlinespace
\multirow{3}{*}{Qwen3-32B}
  & Chinese Homo.    & 13.6\% & 22.7\% & \textbf{54.5\%} & 9.1\% \\
  & Number Homo.     & 0.0\% & 42.9\% & \textbf{57.1\%} & 0.0\% \\
  & Unconv. Chars.   & 50.0\% & \textbf{50.0\%} & 0.0\% & 0.0\% \\
\addlinespace
\multirow{3}{*}{GLM-4-32B}
  & Chinese Homo.    & 27.3\% & \textbf{38.2\%} & 32.7\% & 1.8\% \\
  & Number Homo.     & 0.0\% & \textbf{66.7\%} & 33.3\% & 0.0\% \\
  & Unconv. Chars.   & \textbf{33.3\%} & \textbf{33.3\%} & \textbf{33.3\%} & 0.0\% \\
\addlinespace
\multirow{3}{*}{Gemma3-27B}
  & Chinese Homo.    & 22.2\% & \textbf{44.4\%} & 26.7\% & 6.7\%  \\
  & Number Homo.     & 0.0\% & \textbf{100\%} & 0.0\% & 0.0\% \\
  & Unconv. Chars.   & 0.0\% & 0.0\% & \textbf{100\%} & 0.0\% \\
\bottomrule
\end{tabular}
\caption{Breakdown of Cell B failure modes (Tier 1 correct, Tier 2 wrong) 
across five models and three open-ended subcategories. Each row sums to 100\%. 
\textbf{Copy Verbatim}: model outputs the neologism itself. 
\textbf{Explanation/Paraphrase}: model outputs a semantic gloss or near-synonym 
of the source form. 
\textbf{Wrong Restoration}: model outputs a different Chinese expression 
that is neither the neologism nor a paraphrase. 
\textbf{Empty/Other}: empty output or unparseable.}
\label{tab:failure-mode-breakdown}
\end{table*}
\begin{table*}[t]
\centering
\small
\setlength{\tabcolsep}{5pt}
\renewcommand{\arraystretch}{1.4}
% Column width control: Tier 1 Output gets the most space
\newcolumntype{S}{>{\hsize=0.55\hsize}X}   % Subcategory - narrow
\newcolumntype{M}{>{\hsize=0.55\hsize}X}   % Model - narrow
\newcolumntype{N}{>{\hsize=0.6\hsize}X}    % Neologism - narrow
\newcolumntype{G}{>{\hsize=0.6\hsize}X}    % Gold Source - narrow
\newcolumntype{W}{>{\hsize=1.9\hsize}X}    % Tier 1 Output - wide
\newcolumntype{T}{>{\hsize=0.6\hsize}X}    % Tier 2 Output
\resizebox{0.95\textwidth}{!}{
\begin{tabularx}{\textwidth}{S M N G W T}
\toprule
\textbf{Subcategory} & \textbf{Model} & \textbf{Neologism} & \textbf{Gold} & \textbf{Tier 1 Model Output (correct)} & \textbf{Tier 2 Output (wrong)} \\
\midrule

% ===== Chinese Homo. =====
Chinese Homo. & Kimi-K2.5 & v我50 & 微我50 & 
网络流行语，指通过微信转账50元人民币。源自肯德基``疯狂星期四''促销活动\dots \newline Referring to transferring 50 yuan RMB via WeChat. It originated from KFC's `Crazy Thursday' promotional campaign. & 
微信我50 \newline (additional characters) \\
\addlinespace

Chinese Homo. & Kimi-K2.5 & 石乐志 & 失了智 & 
网络流行语，``失了智''的谐音，指失去理智、做出愚蠢或不理智的行为\dots \newline
A homophonic pun on ``失了智'' (lost one's wits), meaning to lose one's reason and act in a stupid or irrational way.& 
失智 \newline (dementia) \\
\addlinespace

Chinese Homo. & DeepSeek-V3.2 & 河蟹 & 和谐 & 
网络用语中常被用作``删除''或``屏蔽''的委婉说法，因``和谐''与``河蟹''谐音，后者常被用来讽刺内容审查。\newline Often used as a euphemism for `deleting' or `blocking' content. Because 和谐 (harmony) is a homophone of 河蟹 (river crab). & 
删除 \newline (delete) \\
\addlinespace

Chinese Homo. & DeepSeek-V3.2 & 大清药丸 & 大清要完 & 
大清药丸是网络流行语，源自``大清要完''的谐音，实际用来调侃某事物或局势即将崩溃或失败，带有讽刺或幽默意味。\newline A homophonic pun on `大清要完' (the Qing dynasty is finished). It's used to jokingly suggest that something or some situation is on the verge of collapse or failure, with a satirical or humorous undertone. & 
大清朝要完 \newline (additional characters) \\

\midrule
% ===== Number Homo. =====
Number Homo. & Kimi-K2.5 & 3q & thank you & 
英文``thank you''的谐音数字缩写，``3''对应``th''音、``q''对应``you''音。  \newline `3Q' is a homophonic abbreviation of the English `thank you', using numbers and letters: `3' represents the `th' sound and `q' represents the `you' sound. & 谢谢 \newline (Chinese expression for thank you)\\
\addlinespace

Number Homo. & DeepSeek-V3.2 & 666 & 溜溜溜 & 
表示非常厉害、令人惊叹的意思，源自数字6的谐音``溜''。\newline `666' means something is awesome or impressive. It derives from the number 6, a homophone of `溜' (`slick/skilled') & 
牛牛牛 \newline (semantically equivalent) \\
\midrule

% Number Homo. & [Model] & [neologism] & [gold] & 
% [Tier 1 definition\dots] & 
% [Tier 2 output] & 
% Explanation/ Paraphrase \\
% \addlinespace

% Number Homo. & [Model] & [neologism] & [gold] & 
% [Tier 1 definition\dots] & 
% [Tier 2 output] & 
% Explanation/ Paraphrase \\
% \addlinespace

% ===== Unconv. Chars. =====
Unconv. Chars. & Kimi-K2.5 & 占戈哥欠走已 & 战歌起 & 
将``战歌起''三字拆解为偏旁部首（战=占+戈，歌=哥+欠，起=走+己/已）的隐晦写法\dots \newline A cryptic way of writing `战歌起' (cue the battle anthem) by splitting each of the three characters into their radical components. & 
战歌响起 \newline (additional characters) \\
\addlinespace

Unconv. Chars. & Kimi-K2.5 & 口区 & 呕 & 
将汉字``呕''拆分为``口''和``区''二字输入的网络黑话\dots \newline The Chinese character `呕' (to vomit/retch) is broken apart and typed as its two components, `口' and `区'. & 
呕吐 \newline (additional characters) \\
\addlinespace

Unconv. Chars. & DeepSeek-V3.2 & 彳亍口巴 & 行吧 & 
网络用语，由``彳亍''和``口巴''拼接而成。其中``彳亍''是``行''的拆分形式，``口巴''是``吧''的拆分形式。 \newline `彳亍' is the disassembled form of `行' and `口巴' is the disassembled form of `吧'.  & 
好吧 \newline (another expression for okay) \\
\addlinespace

Unconv. Chars. & DeepSeek-V3.2 & 占戈土也 & 战地 & 
``占戈土也''是``战地''的拆分写法\dots \newline `占戈土也' is the disassembled-character way of `战地'(`battlefield') & 
战争 \newline (war) \\

\bottomrule
\end{tabularx}}
\caption{Representative Cell B cases (Tier 1 correct, Tier 2 wrong) 
across the three open-ended subcategories. In each case, the model 
correctly describes the neologism in Tier 1 but, in Tier 2, 
substitutes a semantic equivalent for the source form rather than 
producing the source form itself (Explanation/Paraphrase), which is the dominant failure mode across 
all three subcategories (Table~\ref{tab:failure-mode-breakdown}). English references are provided below the generated definition.}
\label{tab:case-study}
\end{table*}

\clearpage
\subsection{Prompts}
\label{app:prompts}

All prompts used in \textsc{CNeo-Bench} are originally in Chinese. 
For each prompt below, we provide the original Chinese text 
followed by an English translation. The English translations are 
for reference only and were not used in any experiment.

% ============================================
\subsubsection{Task Prompts}
\label{app:task-prompts}

\begin{promptbox}{Tier 1: Definition Generation}
\textbf{Chinese (original):}

\vspace{2pt}
你是一个中文流行文化与语言学专家。请为给定的中文新词/网络流行语生成解释。

要求：

1. 用简洁、准确的中文解释词义（1-2句话）

2. 说明使用场景（可选）

3. 适当的时候可以给出例句（可选）

4. 不要编造不确定的信息

5. 用json格式输出

词语：\{XXX\}

输出格式：

\{"definition": "中文词的解释/定义"\}

\vspace{6pt}
\textbf{English (translation):}

\vspace{2pt}
You are an expert in Chinese popular culture and linguistics. Please generate an explanation for the given Chinese neologism / internet slang term.

Requirements:

1. Use concise and accurate Chinese to explain the meaning (1–2 sentences)

2. Describe the usage scenario when applicable (optional)

3. Provide an example sentence when appropriate (optional)

4. Do not fabricate uncertain information

5. Output in JSON format

Term: \{XXX\}

Output format:

\{"definition": "Explanation/definition of the Chinese term"\}
\end{promptbox}

\begin{promptbox}{Hard Sample Few-shot Definition Generation}
\textbf{Chinese (original):}

\vspace{2pt}
你是一个中文流行文化与语言学专家。请为给定的中文新词/网络流行语生成解释。

要求：

1. 用简洁、准确的中文解释词义（1-2句话）

2. 说明使用场景（可选）

3. 适当的时候可以给出例句（可选）

4. 不要编造不确定的信息

5. 用json格式输出

下面是一些这个新词在真实语境中的使用示例,请结合这些示例推断其含义:

\{examples\_block\}

现在请根据上述示例,为下面的词语生成解释。

词语：\{XXX\}

输出格式：

\{"definition": "中文词的解释/定义"\}

\vspace{6pt}
\textbf{English (translation):}

\vspace{2pt}
You are an expert in Chinese popular culture and linguistics. Please generate an explanation for the given Chinese neologism / internet slang term.

Requirements:

1. Use concise and accurate Chinese to explain the meaning (1–2 sentences)

2. Describe the usage scenario when applicable (optional)

3. Provide an example sentence when appropriate (optional)

4. Do not fabricate uncertain information

5. Output in JSON format

Below are some examples of how this neologism is used in real contexts. Please infer its meaning based on these examples:

\{examples\_block\}

Now, based on the examples above, generate an explanation for the following term.

Term: \{XXX\}

Output format:

\{"definition": "Explanation/definition of the Chinese term"\}
\end{promptbox}

\vspace{6pt}

\begin{promptbox}{Tier 2: Pinyin Abbreviation (Cloze)}
\textbf{Chinese (original):}

\vspace{2pt}
你需要完成一个拼音缩写选择题任务。

任务说明：

给定一句包含(\_\_\_\_)的句子，请从候选选项中选择填入句子后最符合语境的选项。所有候选项都是拼音缩写

要求：

结合句子语境判断最自然、最符合表达习惯的选项

输入：
句子：\{masked\_example\_sentence\}

选项：
\{options\}

请只输出一个字母（A/B/C/D），不要输出其他内容。

\vspace{6pt}
\textbf{English (translation):}

\vspace{2pt}
You need to complete a pinyin abbreviation multiple-choice task.

Task Description:

Given a sentence containing (\_\_\_\_), choose the option that best fits the context after being filled into the sentence. All candidate options are pinyin abbreviations.

Requirements:

Select the most natural and contextually appropriate option based on the sentence.

Input:
Sentence: \{masked\_example\_sentence\}

Options:
\{options\}

Please output only a single letter (A/B/C/D), and do not output any additional content.
\end{promptbox}

\vspace{6pt}

\begin{promptbox}{Tier 2: English Abbreviation (Cloze)}
\textbf{Chinese (original):}
\vspace{2pt}

你需要完成一个英文缩写选择题任务。

任务说明：

给定一句包含(\_\_\_\_)的句子，请从候选选项中选择填入句子后最符合语境的选项。所有候选项都是英文缩写

要求：

结合句子语境判断最自然、最符合表达习惯的选项

输入：
句子：\{masked\_example\_sentence\}

选项：
\{options\}

请只输出一个字母（A/B/C/D），不要输出其他内容。

\vspace{6pt}
\textbf{English (translation):}

\vspace{2pt}
You need to complete an English abbreviation multiple-choice task.

Task Description:

Given a sentence containing (\_\_\_\_), choose the option that best fits the context after being filled into the sentence. All candidate options are English abbreviations.

Requirements:

Select the most natural and contextually appropriate option based on the sentence.

Input:

Sentence: \{masked\_example\_sentence\}

Options:

\{options\}

Please output only a single letter (A/B/C/D), and do not output any additional content.
\end{promptbox}

\begin{promptbox}{Tier 2: Chinese Homo. (Restoration)}
\textbf{Chinese (original):}

\vspace{2pt}
任务说明：

给定一个谐音梗和一句包含该谐音梗的句子，请还原出其最可能的原始表达。

要求：

结合句子语境进行判断

还原后的词语在读音上应与谐音梗高度相似

字数尽量接近

优先选择日常中最常见、最自然的表达

只输出一个最合理的答案

注意：

不要解释

不要输出多个候选

不要输出无关内容

输入：

谐音梗：\{XXX\}

句子：\{example\_sentence\}

请只输出谐音梗的原本表达，不要输出其他内容。

\vspace{6pt}
\textbf{English (translation):}

\vspace{2pt}
Task Description:

Given a homophonic pun and a sentence containing it, recover its most likely original expression.

Requirements:

Determine the answer based on the sentence context

The recovered expression should be highly similar in pronunciation to the homophonic pun

Keep the number of characters as close as possible

Prefer the most common and natural expression used in daily language

Output only the single most reasonable answer

Notes:

Do not provide explanations

Do not output multiple candidates

Do not output unrelated content

Input:

Homophonic pun: \{XXX\}

Sentence: \{example\_sentence\}

Please output only the original expression of the homophonic pun, and do not output any additional content.
\end{promptbox}

\begin{promptbox}{Tier 2: Number Homo. (Restoration)}
\textbf{Chinese (original):}

\vspace{2pt}
任务说明：

给定一个数字谐音梗和一句包含该谐音梗的句子，请还原出其最可能的原始表达。

要求：

结合句子语境进行判断

还原后的词语在读音上应与谐音梗高度相似

字数尽量接近

优先选择日常中最常见、最自然的表达

只输出一个最合理的答案

注意：

不要解释

不要输出多个候选

不要输出无关内容

输入：

谐音梗：\{XXX\}

句子：\{example\_sentence\}

请只输出谐音梗的原本表达，不要输出其他内容。

\vspace{6pt}
\textbf{English (translation):}

\vspace{2pt}
Task Description:

Given a numeric homophonic pun and a sentence containing it, recover its most likely original expression.

Requirements:

Determine the answer based on the sentence context

The recovered expression should be highly similar in pronunciation to the homophonic pun

Keep the number of characters as close as possible

Prefer the most common and natural expression used in daily language

Output only the single most reasonable answer

Notes:

Do not provide explanations

Do not output multiple candidates

Do not output unrelated content

Input:

Homophonic pun: \{XXX\}

Sentence: \{example\_sentence\}

Please output only the original expression of the homophonic pun, and do not output any additional content.
\end{promptbox}

\begin{promptbox}{Tier 2: Lexical Neo. (Cloze)}
\textbf{Chinese (original):}

\vspace{2pt}
请完成中文完形填空任务。

根据句子语境，从选项中选择最合适的一项填入(\_\_\_\_)。

要求：

结合上下文语义

选择最自然、最常见的表达

句子：\{masked\_example\_sentence\}

选项：
\{options\}

请只输出一个字母（A/B/C/D）。

\vspace{6pt}
\textbf{English (translation):}

\vspace{2pt}
Please complete the Chinese cloze task.

Based on the sentence context, choose the most appropriate option to fill in (\_\_\_\_).

Requirements:

Consider the semantic context

Choose the most natural and commonly used expression

Sentence: \{example\_sentence\}

Options:
\{options\}

Please output only a single letter (A/B/C/D).
\end{promptbox}

\begin{promptbox}{Tier 2: Semantic Neo. (Semantic Discrimination)}
\textbf{Chinese (original):}

\vspace{2pt}
你需要完成一个“旧词新用”识别任务。

任务说明：

给定一个词语及其四个使用该词语的句子，其中：

三个句子使用的是该词语的传统含义（常规用法）

一个句子使用的是该词语在网络或当代语境中的新含义（引申义或新用法）

你的任务是找出使用“新含义”的那个句子。

判断标准：

新用法通常不符合该词语的原始、字面或传统语义

通常带有网络语境、隐喻、娱乐化或引申含义

与原始语义存在明显偏离

输入：
词语：\{XXX\}

选项：
\{options\}

请只输出一个字母（A/B/C/D），不要解释。

\vspace{6pt}
\textbf{English (translation):}

\vspace{2pt}
You need to complete a Semantic Discrimination task.

Task Description:

Given a word and four sentences containing that word:

Three sentences use the traditional meaning of the word (conventional usage)

One sentence uses the word with a newer meaning in internet or contemporary contexts (extended or newly emerged usage)

Your task is to identify the sentence that uses the “new meaning.”

Criteria:

The new usage usually does not align with the word’s original, literal, or traditional meaning

It often carries internet-contextual, metaphorical, entertaining, or extended meanings

There is a clear semantic shift from the original meaning

Input:
Word: \{XXX\}

Options:
\{options\}

Please output only a single letter (A/B/C/D), without explanation.
\end{promptbox}

\begin{promptbox}{Tier 2: Unconv. Chars. (Restoration)}
\textbf{Chinese (original):}

\vspace{2pt}
你需要完成一个中文字符还原任务。

任务说明：

给定一个由拆分字、形似字或伪装字符组成的词语，请还原为其原本的标准汉字表达。

要求：

结合字形和常见表达进行还原

输出应为一个自然、常见的中文词语

注意：

只输出还原后的词语

不要解释

不要包含空格或标点

输入：
\{XXX\}

输出：

\vspace{6pt}
\textbf{English (translation):}

\vspace{2pt}
You need to complete a Chinese character restoration task.

Task Description:

Given a word composed of split characters, visually similar characters, or disguised characters, restore it to its original standard Chinese expression.

Requirements:

Restore the expression based on character shape and common usage

The output should be a natural and commonly used Chinese word or phrase

Notes:

Output only the restored expression

Do not provide explanations

Do not include spaces or punctuation

Input:
\{XXX\}

Output:
\end{promptbox}

\begin{promptbox}{Tier 2: Formulaic. (Scenario Matching)}
\textbf{Chinese (original):}

\vspace{2pt}
你需要完成一个“情境表达选择”任务。

任务说明：

给定一个具体场景和一个问题，请从四个选项中选择最符合该问题的一句话。

判断标准：

是否符合当前情境中的人物行为和氛围

是否自然、符合日常或网络表达习惯

优先选择最贴切，而不是字面最接近的

输入：

场景：

\{scenario\}

问题：

在这个情景下，最符合以下哪个梗？

选项：

\{options\}

请只输出一个字母（A/B/C/D），不要解释。

\vspace{6pt}
\textbf{English (translation):}
\vspace{2pt}

You need to complete a “Scenario Matching” task.

Task Description:

Given a specific scenario and a question, choose the sentence that best fits the question from four options.

Criteria:

Whether it matches the characters’ behavior and the atmosphere in the current scenario

Whether it sounds natural and aligns with everyday or internet language usage

Prioritize the most contextually appropriate choice rather than the most literal one

Input:

Scenario:

\{scenario\}

Question:

Which sentence best fits this scenario?

Options:

\{options\}

Please output only a single letter (A/B/C/D), without explanation.

\end{promptbox}

\begin{promptbox}{Tier 2: Foreign (Source Identification)}
\textbf{Chinese (original):}

\vspace{2pt}
你需要完成一个“流行语来源溯源”任务。

任务说明：

给定一个网络流行语及其使用语境，请从四个选项中选择最符合其真实来源和形成逻辑的解释。

输入：

问题：

网络流行语 \{XXX\} 产生的根本原因/来源逻辑是？

选项：
\{options\}

请只输出一个字母（A/B/C/D），不要解释。

\vspace{6pt}
\textbf{English (translation):}

\vspace{2pt}
You need to complete an “source identification” task.

Task Description:

Given an internet slang term and its usage context, choose the explanation that best matches its true origin and formation logic from four options.

Input:

Question:

What is the fundamental reason / source logic behind the emergence of the internet slang term \{XXX\}?

Options:
\{options\}

Please output only a single letter (A/B/C/D), without explanation.
\end{promptbox}

% ... 重复上面的 block,填入剩余 7 个 Tier 2 subcategory:
% Chinese Homo. (Restoration)
% Number Homo. (Restoration)
% Unconv. Chars. (Restoration)
% Lexical Neo. (Cloze)
% Semantic Neo. (Sentence Selection)
% Formulaic (Scenario Matching)
% Foreign (Source Attribution)

% ============================================
\subsubsection{Data Construction Prompts}
\label{app:construction-prompts}

These prompts were used during dataset construction to generate 
content with an LLM. For each item lacking an original example sentence in the source 
materials, we use GPT-5.4 to 
generate an example sentence. For cloze tasks, we use the same 
model to generate distractor options; for scenario-matching tasks 
(Formulaic.), we use it to generate the scenario text. All 
LLM-generated contents are subsequently reviewed and corrected by 
the authors to ensure quality and naturalness.

\begin{promptbox}{Example Sentence Generation}
\textbf{Chinese (original):}

\vspace{2pt}
你是一个中文流行文化与语言学专家。请根据提供的中文新词及其定义，构造数据。

任务要求：

1. **语境例句生成**：编写一个自然、逻辑通顺的中文句子，句子中必须包含给出的“中文新词”，注意**不要出现它的提示或解释**。

2. **输出限制**：必须以严格的 JSON 格式输出，不得包含任何解释或 Markdown 标签。

输入数据：

- 中文新词：\{XXX\}

- 定义解释：\{definition\}

输出格式：

\{"example": "例句内容"\}

\vspace{6pt}
\textbf{English (translation):}

\vspace{2pt}
You are an expert in Chinese popular culture and linguistics. Please construct data based on the provided Chinese neologism and its definition.

Task Requirements:

1. Contextual Example Generation: Write a natural and logically coherent Chinese sentence that must contain the given “Chinese neologism.” Note that the sentence should not include any hints or explanations of the term.

2. Output Restriction: The output must be in strict JSON format, without any explanations or Markdown tags.

Input Data:

- Chinese Neologism: \{XXX\}

- Definition: \{definition\}

Output Format:

\{"example": "example sentence"\}
\end{promptbox}

\vspace{6pt}

\begin{promptbox}{Cloze Generation}
\textbf{Chinese (original):}

\vspace{2pt}
你是一个中文语言测评专家。请根据提供的“中文新词”及其例句和定义，构造一个高质量的完形填空（Cloze）测试题。

任务步骤：

1. **设置干扰项**：

    - **正确项**："\{word\}"。
    
    - **干扰项 (Distractors)**：生成 3 个与“\{word\}”**词性相同**（如都是名词或形容词）且**代入例句后语法完全正确**的词语。
    
    - **迷惑性要求**：干扰项在字面上应看起来属于同一领域，但在该语境下的逻辑准确度不如“\{word\}”, 注意不要出现“\{word\}”的同义词。
    
2. **构造题目**：将例句中的“\{word\}”替换为 `(\_\_\_\_)`。

输入数据：

- 目标新词：\{word\}

- 新词例句：\{example\_sentence\}

- 词义参考：\{definition\}

输出格式（严格 JSON）：

\{
  "cloze\_question": "带有 (\_\_\_\_) 的句子",
  "options": \{"A": "...", "B": "...", "C": "...", "D": "..."\},
  "answer": "正确选项的字母"
\}

\vspace{6pt}
\textbf{English (translation):}

\vspace{2pt}
You are an expert in Chinese language evaluation. Based on the provided “Chinese neologism,” its example sentence, and its definition, construct a high-quality cloze test question.

Task Steps:

1. Set Distractors:
- Correct Option: "\{word\}".

- Distractors: Generate 3 words that have the same part of speech as “\{word\}” (e.g., all nouns or all adjectives) and are grammatically valid when inserted into the example sentence.

- Plausibility Requirement: The distractors should appear to belong to the same general domain on the surface, but their semantic fit in the given context should be less accurate than “\{word\}”. Do not use synonyms of “\{word\}”.

2. Construct the Question:
Replace “\{word\}” in the example sentence with (\_\_\_\_).

Input Data:

- Target Neologism: \{word\}

- Example Sentence: \{example\_sentence\}

- Reference Definition: \{definition\}

Output Format (strict JSON):

\{
"cloze\_question": "Sentence containing (\_\_\_\_)",
"options": {"A": "...", "B": "...", "C": "...", "D": "..."},
"answer": "Letter of the correct option"
\}
\end{promptbox}

\vspace{6pt}

\begin{promptbox}{Scenario Generation}

\textbf{Chinese (original):}
\vspace{2pt}

你是一个精通中文社会语境和网络亚文化的专家。请根据提供的“梗”及其定义，构造一个考察“语境应用能力”的情景。

任务核心：

设计通过一个复杂的情境，测试模型是否理解该梗的使用动机和情绪色彩。

题目构造要求：

结合定义编写一个 50-100 字的故事背景。该情景需要严格符合目标词的含义，但不能出现目标词的解释或者提示。

输入：

- 目标词：\{word\}

- 含义参考：\{definition\}

输出格式（严格 JSON）：

\{
  "scenario": "情景" \}

\vspace{6pt}
\textbf{English (translation):}
\vspace{2pt}

You are an expert in Chinese social context and internet subcultures. Based on the provided meme/slang term and its definition, construct a scenario that tests contextual application ability.

Core Task:

Design a complex scenario to test whether the model understands the usage motivation and emotional tone of the meme/slang term.

Question Construction Requirements:

Write a 50–100 character story background based on the definition. The scenario must strictly align with the meaning of the target term, but must not include the explanation or explicit hints about the term.

Input:

- Target Term: \{word\}
- Reference Definition: \{definition\}

Output Format (strict JSON):

\{
"scenario": "Scenario"
\}

\end{promptbox}

\vspace{6pt}

\begin{promptbox}{Distractor Generation (Source Identification)}
\textbf{Chinese (original):}

\vspace{2pt}
你是一个负责设计高质量选择题的语言学专家。你的任务是为一个'中文外来语溯源'选择题生成 3 个干扰项（错误选项），使得这道题能够真正测试模型对被测词的知识，而不是让模型靠表面特征蒙对。

输入信息:

- **被测词**：\{chinese\}

- **正确来源**：\{correct\_origin\}

生成规则:

1. 结构平行规则

所有 4 个选项必须遵循与给出的正确来源完全相同的句式模板:
不允许有任何结构变体。不允许某个选项多出一个分句或少一个分句。

2. 语言同源规则

所有 4 个选项必须**来自同一种源语言**(如果正确答案是英文,那么所有干扰项也必须是英文音译;如果是日语,则全部是日语音译)。不允许混入其他语言作为干扰项，这种"语言错配"型干扰太容易被模型识别。

3. 场景同源规则

所有 4 个选项的流行场景必须属于**同一个大类**(比如都和游戏有关,或都和动漫有关,或都和社交媒体有关)。场景差异过大会让模型靠常识排除。

4. 可信度规则

每个干扰项本身都应该是"听起来合理的词源解释"，一个不知道真相的人类读者应该**无法仅凭常识排除**任何一个干扰项。如果某个干扰项一眼就能看出是假的,它就没有起到干扰作用,必须重写。

5. 输出格式（严格 JSON）

\vspace{6pt}
\textbf{English (translation):}

\vspace{2pt}
You are a linguistics expert responsible for designing high-quality multiple-choice questions. Your task is to generate 3 distractors (incorrect options) for a “Chinese loanword origin tracing” multiple-choice question, so that the question genuinely tests the model’s knowledge of the target word rather than allowing it to guess based on superficial patterns.

Input Information:

Target Word: \{chinese\}

Correct Origin: \{correct\_origin\}

Generation Rules:

1. Structural Parallelism Rule

All 4 options must follow exactly the same sentence template as the given correct origin.
No structural variations are allowed. No option may contain an extra clause or omit a clause.

2. Same Source Language Rule

All 4 options must come from the same source language.
(For example, if the correct answer is an English-derived transliteration, then all distractors must also be English-derived transliterations; if it is Japanese-derived, then all distractors must also be Japanese-derived.)
Do not mix in other languages as distractors, since this type of “language mismatch” is too easy for models to detect.

3. Same Scenario Rule

All 4 options must belong to the same broad usage domain (e.g., all related to gaming, all related to anime, or all related to social media).
If the domains differ too much, the model can eliminate options using common sense alone.

4. Plausibility Rule

Each distractor should itself sound like a “plausible etymological explanation.”
A human reader who does not know the true answer should not be able to eliminate any distractor based only on common sense.
If a distractor is obviously fake at a glance, it fails as a distractor and must be rewritten.

5. Output Format (strict JSON)
\end{promptbox}

% ... 其余 construction prompts(如 Semantic Neo. 句子生成、
% Formulaic scenario 生成等),按同样的 block 复制

% ============================================
\subsubsection{Judge Prompt}
\label{app:judge-prompt}

This prompt is used by the LLM judge (Qwen3-235B-A22B-Instruct-2507) 
to score Tier 1 definition generation outputs.

\begin{promptbox}{Tier 1 Definition Generation: LLM-as-Judge}
\textbf{Chinese (original):}

\vspace{2pt}
你是一个语言学评估助手，需要判断"模型生成的定义"是否与"标准答案"语义一致。

评估标准：

- 1（正确）：核心含义一致，即使表达方式不同、略有简化或举例不同，也算正确

- 0（错误）：核心含义不一致、理解错误、缺失关键含义，或完全不相关

注意：

- 判断"语义是否一致"

- 如果定义只覆盖部分含义，但不影响核心理解，可以判为1

- 如果存在明显误解或方向错误，必须判为0

- 如果只有部分正确，其他部分出现明显错误，也需要判为0

- 如果模型答案是空字符串，也需要判为0

- 不要受措辞风格影响

输入：

词语：\{word\}

标准答案：
\{reference\}

模型答案：
\{prediction\}

只要输出一个数字：
0 或 1

\vspace{6pt}
\textbf{English (translation):}

\vspace{2pt}
You are a linguistic evaluation assistant. Your task is to determine whether the “model-generated definition” is semantically consistent with the “reference answer”.

Evaluation Criteria:

- 1 (Correct): The core meaning is consistent, even if the wording differs, is slightly simplified, or uses different examples

- 0 (Incorrect): The core meaning is inconsistent, misunderstood, missing key meaning, or completely irrelevant

Notes:

- Judge whether the semantics are consistent

- If the definition only covers part of the meaning but does not affect the core understanding, it can still be judged as 1

- If there is an obvious misunderstanding or incorrect direction, it must be judged as 0

- If only part is correct but other parts contain clear errors, it should also be judged as 0

- If the model answer is an empty string, it should also be judged as 0
Do not be influenced by wording or writing style

Input:

Word: \{word\}

Reference Answer:
\{reference\}

Model Answer:
\{prediction\}

Output only a single number:
0 or 1
\end{promptbox}
\end{CJK*}
\end{document}